\documentclass{article}

\usepackage[preprint]{neurips_2025}

\usepackage[utf8]{inputenc} % allow utf-8 input
\usepackage[T1]{fontenc}    % use 8-bit T1 fonts
\usepackage{hyperref}       % hyperlinks
\usepackage{url}            % simple URL typesetting
\usepackage{booktabs}       % professional-quality tables
\usepackage{amsfonts}       % blackboard math symbols
\usepackage{nicefrac}       % compact symbols for 1/2, etc.
\usepackage{microtype}      % microtypography
\usepackage{pifont}

\usepackage{wrapfig}
\usepackage{multirow}
\usepackage{multicol}
\usepackage{arydshln} % for dashed lines in tables
\usepackage{graphicx}
\usepackage{array}
\usepackage[table,xcdraw]{xcolor}
\usepackage{caption}
\usepackage{colortbl}
\usepackage{xparse}
\usepackage{graphicx}
\definecolor{mygreen}{RGB}{34,139,34}
\usepackage{wrapfig}
\usepackage{lipsum}
\usepackage{amsmath}
\definecolor{tablehighlightgray}{RGB}{231, 231, 231}

\newcommand{\eg}{\textit{e}.\textit{g}.}

\newcommand{\etc}{\textit{etc}}

\title{\textcolor{red}{L}\textcolor{orange}{e}\textcolor{yellow!80!black}{a}\textcolor{teal}{d}\textcolor{cyan}{e}\textcolor{blue}{r}\textcolor{violet}{360}\textcolor{brown}{V}:
A \textcolor{red}{L}arge-scale,
R\textcolor{orange}{e}al-world \textcolor{violet}{360} \textcolor{brown}{V}ideo Dataset for
Multi-t\textcolor{yellow!80!black}{a}sk Learning in
\textcolor{teal}{D}iverse
\textcolor{cyan}{E}nvi\textcolor{blue}{r}onments
}

\author{Weiming Zhang$^{1*}$ \quad Dingwen Xiao$^{1*}$ \quad Aobotao Dai$^{1*}$ \quad Yexin Liu$^{2}$  \\ \quad \textbf{Tianbo Pan$^{3}$}  \quad \textbf{Shiqi Wen$^{1}$} \quad \textbf{Lei Chen$^{1,2}$} \quad \textbf{Lin Wang$^{4\dag}$ }
\vspace{1mm}\\
$^{1}$ HKUST (GZ) \quad $^{2}$ HKUST \quad $^{3}$ National University of Singapore  \quad $^{4}$ Nanyang Technological University
\vspace{1mm}\\}

\begin{document}

\maketitle

\begin{figure}[h!]
    \centering
    \vspace{-25pt}
\includegraphics[width=0.9\linewidth]{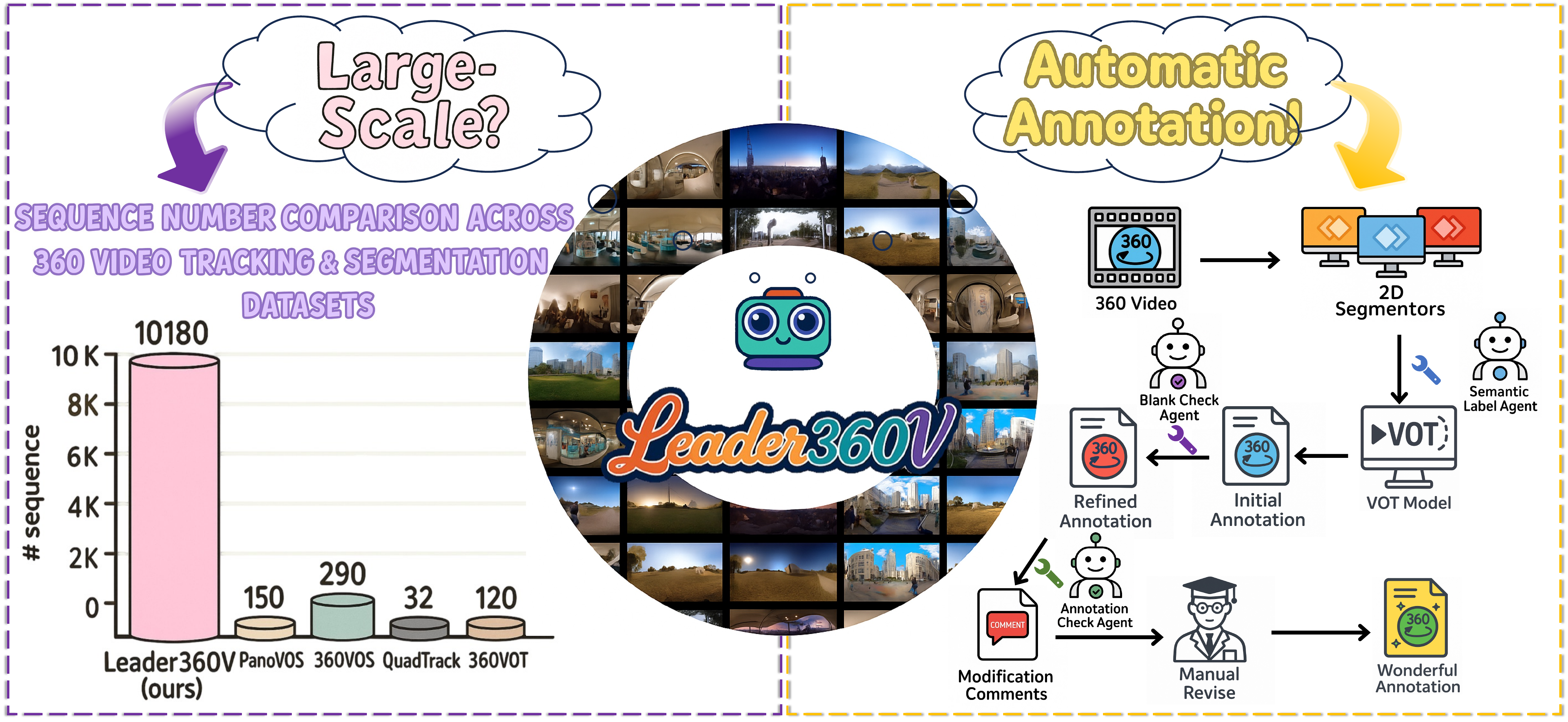}
\caption{\small{The overall of our Leader360V dataset.}}
\vspace{-10pt}
\label{fig: teaser_figure}
\end{figure}

% updated by addison
\begin{abstract}
% 360 video offers holistic environmental perception, making it particularly valuable for applications in autonomous driving, robotics, and immersive scene understanding.
360 video captures the complete surrounding scenes with the ultra-large field of view of 360$\times$180. This makes 360 scene understanding tasks, \eg, segmentation and tracking, crucial for appications, such as autonomous driving, robotics. With the recent emergence of foundation models, the community is, however, impeded by the lack of large-scale, labelled real-world datasets. This is caused by the inherent spherical properties, \eg, severe distortion in polar regions, and content discontinuities, rendering the annotation costly yet complex.
% However, the unique characteristics of 360-degree video—such as a wide field of view, severe geometric distortion, and content continuity across horizontal borders—pose significant challenges for annotation. These issues have substantially constrained the scale and quality of existing datasets for segmentation and tracking.
% However, the inherent geometric properties of 360 video—wide field of view, severe distortion, and horizontal continuity—significantly increase the cost and complexity of manual annotation, limiting the scalability of segmentation and tracking video datasets.
This paper introduces \textbf{Leader360V}, the \textbf{first} large-scale (10K+), labeled real-world 360 video datasets for instance segmentation and tracking.
Our datasets enjoy high scene diversity, ranging from indoor and urban settings to natural and dynamic outdoor scenes. To automate annotation, we design an automatic labeling pipeline, which subtly coordinates pre-trained 2D segmentors and large language models (LLMs) to facilitate the labeling. The pipeline operates in three novel stages. Specifically, in the \textbf{Initial Annotation Phase}, we introduce a Semantic- and Distortion-aware Refinement (\textbf{SDR}) module, which combines object mask proposals from multiple 2D segmentors with LLM-verified semantic labels. These are then converted into mask prompts to guide SAM2 in generating distortion-aware masks for subsequent frames. In the \textbf{Auto-Refine Annotation Phase}, missing or incomplete regions are corrected either by applying the SDR again or resolving the discontinuities near the horizontal borders. The \textbf{Manual Revision Phase} finally incorporates LLMs and human annotators to further refine and validate the annotations.
% for better mask quality and consistency.
Extensive user studies and evaluations demonstrate the effectiveness of our labeling pipeline. Meanwhile, experiments confirm that Leader360V significantly enhances model performance for 360 video segmentation and tracking, paving the way for more scalable 360 scene understanding.

\end{abstract}

\vspace{-10pt}
\section{Introduction}
\label{sec:Introduction}
% Panoramic 360 video provides an omnidirectional view of the environment, offering rich visual context invaluable for applications such as autonomous driving, robotics, and immersive virtual reality. By capturing the full surrounding scene, 360 video enables comprehensive situational awareness that surpasses the field-of-view constraints of conventional cameras. A widely adopted representation format for such data is the equirectangular projection (ERP), which flattens the spherical view into a rectangular image. While popular and compatible with standard vision models, ERP introduces unique geometric challenges.
360 cameras, \textit{a.k.a}, panoramic cameras, provide an ultra-large field of view (FoV) of 360$\times$180 for the surrounding environment. Therefore, 360 video enables comprehensive situational awareness that surpasses the FoV limitations of perspective 2D cameras. This makes 360 camera-based scene understanding popular and crucial for applications such as autonomous driving~\cite{wen2024panacea, zhang2022openmpd, petrovai2022semantic, zhang2024goodsam, Yan2023PanoVOSBN}, robotics~\cite{zhang2018detection,huang2022360vo}, and virtual reality~\cite{chen2024360+, chang2022omniscribe}. A commonly used representation for 360 videos is the equirectangular projection (ERP), which maps the spherical content onto a 2D rectangular plane to ensure compatibility with the standard imaging pipeline.
However, ERP poses several challenges specific to 360 video, including projection distortions in polar regions, and horizontal discontinuities~\cite{Yan2023PanoVOSBN} that break content continuity across the left and right borders. These challenges significantly increase the cost and complexity of manual annotation for 360 videos.

Although several 360 video benchmarks~\cite{Yan2023PanoVOSBN, Xu2024360VOTSVO, Luo2025OmnidirectionalMT} have been proposed for scene understanding tasks, such as segmentation and tracking, the scale and diversity of these datasets remain limited by far in the community, especially with the recent emergence of foundation models \cite{Ravi2024SAM2S}. For segmentation, 360VOS~\cite{Xu2024360VOTSVO} contains 290 panoramic sequences annotated across 62 categories, while PanoVOS~\cite{Yan2023PanoVOSBN} provides 150 high-resolution videos with instance masks. For the tracking task, 360VOT~\cite{Huang2023360VOTAN} focuses on single-object tracking with 120 omnidirectional videos covering 32 object types, and QuadTrack~\cite{Luo2025OmnidirectionalMT} introduces a small-scale multi-object tracking benchmark under non-uniform motion. However, the modest size and task-specific design of these datasets limit their ability to support large-scale, generalizable model training. In contrast, recent 2D video benchmarks such as YouTube-VOS~\cite{Xu2018YouTubeVOSAL}, with over 3,400 videos and 540K segmentation annotations, and SA-V~\cite{Ravi2024SAM2S}, with 50.9K videos and over 640K masklets, have demonstrated the value of dense, large-scale annotation for training foundation models.

\begin{wrapfigure}{r}{0.5\linewidth}
\vspace{-15pt}
  \begin{center}
    \includegraphics[width=0.5\textwidth]{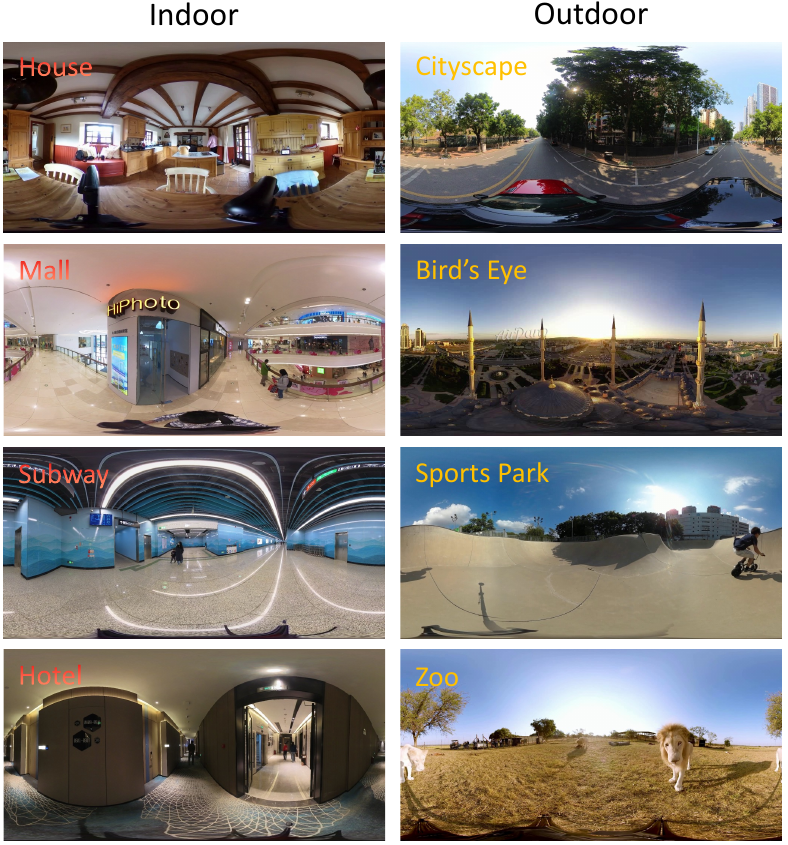}
  \end{center}
  \vspace{-5pt}
  \caption{\small{Samples from different scenarios of the Leader360V dataset.}}
\vspace{-15pt}
\label{fig: dataset_sample}
\end{wrapfigure}

This disparity raises a key scientific question: \textit{Can we build a large-scale 360 video dataset with rich annotations for both segmentation and tracking tasks, while substantially reducing the human labeling cost?}
In this paper, we present \textbf{Leader360V} (Sec.~\ref{sec: Leader360V dataset}), the first large-scale (10K+), real-world 360 video dataset with dense, frame-level annotations for scene understanding tasks across segmentation and tracking. Leader360V covers 198 object types and covers a wide variety of scenes, including both indoor and outdoor environments, as shown in Fig.~\ref{fig: dataset_sample}. Leader360V is constructed by integrating existing public datasets with our self-collected 360 videos captured in diverse real-world environments, yielding a scalable and representative benchmark for panoramic understanding.

To enable the construction of Leader360V, we also propose A$^3$360V (Automatic Annotate Any 360 Video)(Sec. \ref{sec: Automatic Annotation Pipline}), a novel annotation pipeline tailored for 360 videos. A$^3$360V is designed to reduce manual labeling burden while maintaining high annotation quality through a three-phase pipeline:
\textbf{Initial Annotation Phase}(Sec. \ref{sec: Initial Annotation Phase}): We first extract keyframes and use multiple 2D segmentors (e.g., CropFormer~\cite{qi2022high}, OneFormer~\cite{jain2023oneformer}) to generate semantic and instance segmentation proposals. These outputs are unified and aligned via LLM-based semantic matching, then refined through a Semantic- and Distortion-aware Refinement (SDR) Module that leverages SAM2 to produce high-quality panoramic masks.
\textbf{Auto-Refine Annotation Phase}(Sec. \ref{sec: Auto-Refine Annotation Phase}): For subsequent keyframes in the video, we iteratively propagate annotations and identify low-quality regions based on mask coverage. Frames failing coverage thresholds are reprocessed using a GPT-guided Motion-Continuity Refinement (MCR) module, which resolves annotation inconsistencies across left-right ERP and recovers missing masks caused by occlusion or distortion.
\textbf{Manual Revise Phase}(Sec. \ref{sec: Manual Revise Phase}): Finally, human annotators validate and correct the outputs from the previous stages. Multi-annotator review ensures consistency and completeness across frames, producing the final high-quality annotations.

Extensive validation confirms the effectiveness of our pipeline. User studies show that A$^3$360V significantly reduces annotator workload while preserving annotation quality. Experiments on standard 360 video segmentation and tracking benchmarks demonstrate that Leader360V enhances model performance, paving the way for robust, scalable, and generalizable 360 video understanding.

In summary, our contributions are three-fold: (\textbf{I}) We propose \textbf{Leader360V}, the first large-scale (10K+), labeled real-world 360 video dataset specifically designed for instance segmentation and tracking in diverse and dynamic environments. (\textbf{II}) We also propose \textbf{A$^3$360V} (Automatic Annotate Any 360 Video) pipeline, which integrates pre-trained 2D segmentors with large language models to automate the annotation process and significantly reduce human effort without compromising label quality.
 (\textbf{III}) Extensive user studies and experimental results validate the effectiveness of our Lead360V and proposed pipeline and highlight the potential of Leader360V to advance robust 360 video understanding.

\begin{table*}[t!]
\centering
{\caption
{\small \textbf{Comparison of 360 video datasets on segmentation (VOS) and tracking (VOT)}.  \textbf{``Mobile''}: videos shot with motion. \textbf{``Still''}: videos shot without any motion. \textbf{``Vehicle''}: videos shot on vehicles. \textbf{``Human''}: videos shot by humans while walking or running.  \textbf{``Attr''}: characteristics of tracking (Single-Object and Multi-Object Tracking) and segmentation (Partial Frame and Whole Frame Segmentation). \textbf{``Auto''}: no human involvement except revise. \textbf{``Manu''}: no assistant model involvement.}
% {\textbf{Comparison of 360 video datasets on segmentation (VOS) and tracking (VOT)}. \textbf{``Vol''} refers to the size of the dataset. \textbf{``State''} indicates how the videos were shot, ``Mobile'' indicates how many videos were shot with motion, and ``Still'' indicates how many videos were shot without any motion. \textbf{``Foundation''} indicates how the mobile videos were shot, and ``Vehicle'' indicates how many videos were shot on some kind of vehicle, including cars, drones, etc. ``Human'' indicates how many videos were shot by humans while walking or running. \textbf{``Avg''} refers to the average video length, \textbf{``Num''} indicates the total number of categories, \textbf{``Attr''} describes the characteristics of tracking and segmentation. For tracking, there are two types: SOT (single object tracking) and MOT (Multi-Object Tacking). For segmentation, there are two types: PFS (Partial Frame Segmentation) and WFS (Whole Frame Segmentation). \textbf{``Anno''} refers to the annotation methods used. ``Anto'' means no human involvement during annotation, ``Semi'' means additional models assist the human annotator, and ``Manu'' means no assistant model is involved during annotation.}
\label{table:comparison_360video}}
\setlength{\tabcolsep}{0.4mm}
\small
   \begin{tabular}{crcccccccccccc}
        \toprule

        \multirow{2}{*}{\bf{Task}} & \multirow{2}{*}{\bf{Dataset}} & \multirow{2}{*}{\bf{Vol}} & & \multicolumn{2}{c}{\bf{State}} & & \multicolumn{2}{c}{\bf{Foundation}} & & \multirow{2}{*}{\bf{Avg}} & \multirow{2}{*}{\bf{Class}} & \multirow{2}{*}{\bf{Attr}} & \multirow{2}{*}{\bf{Anno}} \\
        % \cmidrule{2-4}
        \cmidrule{5-6}
        \cmidrule{8-9}
        & & & & Mobile & Still & & Vehicle & Human & & &  &
        \\
        \midrule

        \multirow{5}{*}{360VOT} & 360VOT \cite{Xu2024360VOTSVO} & 120 & & 96 & 24 & & 89 & 7 & & 940f  & 32 & SOT & Semi \\

         & PanoVOS \cite{Yan2023PanoVOSBN} & 150 & & 21 & 129 & & 13 & 8 & & 20s  & 35 &  SOT & Semi \\

        & QuadTrack \cite{Luo2025OmnidirectionalMT} & 32 & &
        32 & 0 & & 32 & 0 & & 60s & N/A & MOT & Manual \\

        & JRDB \cite{MartnMartn2021JRDBAD} & 54 & & 32 & 22 & & 32 & 0 & & 70s & N/A & MOT & Manual\\

        \rowcolor{tablehighlightgray}
        \cellcolor{white} & Leader360-T (Ours) & 10180 & & 5K+ & 5K+  & & 2K+ & 3K+ &   & 15s & 198 & MOT & Auto \\

        \midrule

        \multirow{5}{*}{360VOS} & 360VOS \cite{Xu2024360VOTSVO} & 170 & & 135 & 35 & & 124 & 11 & & 940f  & 32 & PFS & Semi \\

         & PanoVOS \cite{Yan2023PanoVOSBN} & 150 & &21 & 129 & & 13 & 8 &  & 20s  & 35 &  PFS & Semi\\

        & WOD \cite{Mei2022WaymoOD} & 1150 & & 1150 & 0 & & 1150 & 0 &  & 20s & 28 & WFS & Manual\\

        \rowcolor{tablehighlightgray}
        \cellcolor{white} & Leader360-S (Ours) & 10180 & & 5K+ & 5K+ & & 2K+ & 3K+ &   & 15s & 198 & WFS & Auto \\

        \bottomrule
   \end{tabular}
   % % \vspace{20pt}
   % \parbox{\linewidth}{\vspace{5pt}
   % \centering
   %      % \footnotesize
   %      \small

   %  }
   %  \parbox{\linewidth}
    \vspace{-15pt}

\end{table*}

\section{Related Works}
\label{sec:related_work}
\textbf{Video-based panoramic datasets for object tracking and segmentation.} 360 video, with its omnidirectional coverage, offers advantages over conventional 2D video, such as a broader field of view, richer spatial context, and greater understanding of the continuous scene. These benefits have led to the development of various 360 video datasets across different tasks, including object tracking~\cite{Huang2023360VOTAN, Luo2025OmnidirectionalMT, MartnMartn2021JRDBAD}, and segmentation~\cite{Yan2023PanoVOSBN, Xu2024360VOTSVO, Mei2022WaymoOD}. Object tracking in 360 videos has been explored through single-object and multi-object tracking benchmarks. For instance, 360VOT~\cite{Huang2023360VOTAN} provides the first dataset for omnidirectional single-object tracking, while QuadTrack~\cite{Luo2025OmnidirectionalMT} captures non-uniform motion using a quadruped robot to establish a multi-object tracking challenge. Segmentation, which is more annotation-intensive, demands pixel-level masks and is mainly represented by datasets focused on instance and panoptic segmentation, such as PanoVOS~\cite{Yan2023PanoVOSBN} and 360VOS~\cite{Xu2024360VOTSVO}. These datasets help address 360-specific challenges such as distortion and content continuity. However, most existing datasets remain limited in scale and task diversity, restricting their ability to support robust and generalizable learning. \textit{To this end, we introduce Leader360V, a large-scale 360 video dataset constructed by integrating publicly available resources and newly self-collected videos, enhanced by an automatic annotation pipeline for multi-task learning. Detailed comparison is shown in Tab. \ref{table:comparison_360video}.}

\textbf{Automated Annotation Frameworks for Scalable Dataset Construction.}
As large-scale video datasets continue to grow, the demand for efficient annotation has led to the emergence of semi-automatic and automatic pipelines aimed at reducing manual labeling costs. For the annotation of 360 video, it is a complex task that necessitates specialized attention due to its unique characteristics, such as severe distortion, a wide field of view, and discontinuous context across panoramic borders. 360 video annotation methods such as 360Rank~\cite{Song2025FineGrainedPI} and PanoVOS~\cite{Yan2023PanoVOSBN} adopt semi-supervised pipelines using pre-trained segmentors and keyframe propagation, but still rely heavily on manual mask drawing and semantic labeling, limiting scalability in 360 settings. However, these methods are not essentially different from 2D video annotation strategies~\cite{Ravi2024SAM2S, Hong2022LVOSAB, Miao2022LargescaleVP} and do not take into account the special characteristics of 360 videos. To address these gaps, we learn from recent automatic annotation systems~\cite{Zhou2024OpenAnnotate2MA, Zhou2024ALGPTMC}, which have incorporated large language models (LLMs) to further reduce human involvement. \textit{We propose A$^3$360V, a unified annotation framework tailored for 360 videos. By integrating LLMs for semantic role assignment and pre-trained 2D segmentors for initial mask generation, A$^3$360V enables scalable segmentation and tracking from keyframes to full video sequences under omnidirectional conditions.}

\textbf{Large‑Scale 2D Video Segmentation and Tracking Datasets.}
Compared to the challenges faced in constructing large-scale 360 video datasets, the field of 2D video understanding has witnessed the emergence of numerous large-scale datasets for segmentation and tracking tasks. YouTube-VOS~\cite{Xu2018YouTubeVOSAL}, LVOS~\cite{Hong2022LVOSAB}, MeViS~\cite{Ding2023MeViSAL}, VIPSeg~\cite{Miao2022LargescaleVP}, SA-V~\cite{Ravi2024SAM2S} for segmentation task and TrackingNet~\cite{Mller2018TrackingNetAL}, LaSOT~\cite{Fan2018LaSOTAH}, TAO~\cite{Dave2020TAOAL} for tracking task have a large base or long shots of a single video, the largest of which can exceed 50K, and a single video can exceed 7 hours.
\textit{Motivated by the gap between the rapid expansion of 2D video resources and the limited availability of large-scale 360 video datasets, we introduce Leader360V, a richly annotated 360 video dataset designed for segmentation and tracking tasks. }

\begin{wraptable}{r} {0.45\linewidth}
    \centering
    \vspace{-25pt}
    \caption{\small{\textbf{Our Data Source}.  \textbf{``Pct''}: percentage of selected data. \textbf{``Sel''}: specific number of selected data. VG: Video Generation. VC: Video Caption}}
   \label{table:datasource}
    \vspace{-5pt}
    \small
    \setlength{\tabcolsep}{0.4mm}
   \begin{tabular}{rcccc}
        \toprule

        {\bf{Source*}} & {\bf{Task}} &  {\bf{Pct}} & {\bf{Sel}} & {\bf{Relabel}}\\
        % \cmidrule{2-4}
        \midrule

         360VOTS* \cite{Xu2024360VOTSVO} & 360VOT & 80\% & 232 & \textcolor{mygreen}{\ding{51}} \\
         PanoVOS* \cite{Yan2023PanoVOSBN} & 360VOS & 60\% & 90 & \textcolor{mygreen}{\ding{51}} \\
         WEB360* \cite{wang2024360dvd}  & 360 VG & 50\% & 1K+ & \textcolor{mygreen}{\ding{51}}\\
         360+x* \cite{chen2024360+} & 360 VC & 30\% & 1K+ & \textcolor{mygreen}{\ding{51}} \\
         YouTube360* \cite{tan2024imagine360} & 360 VC & 20\% & 3K+ & \textcolor{mygreen}{\ding{51}}\\
         Open Source & N/A & N/A & 1K+ & \textcolor{mygreen}{\ding{51}} \\
         Self-Collected & 360VOTS & N/A & 2K+ & \textcolor{mygreen}{\ding{51}}\\
        % 1057 646 1911
        \bottomrule
        % \vspace{-10pt}
   \end{tabular}
\end{wraptable}

\begin{wrapfigure}{r}{0.35\linewidth}
\vspace{-35pt}
  \begin{center}
    \includegraphics[width=0.35\textwidth]{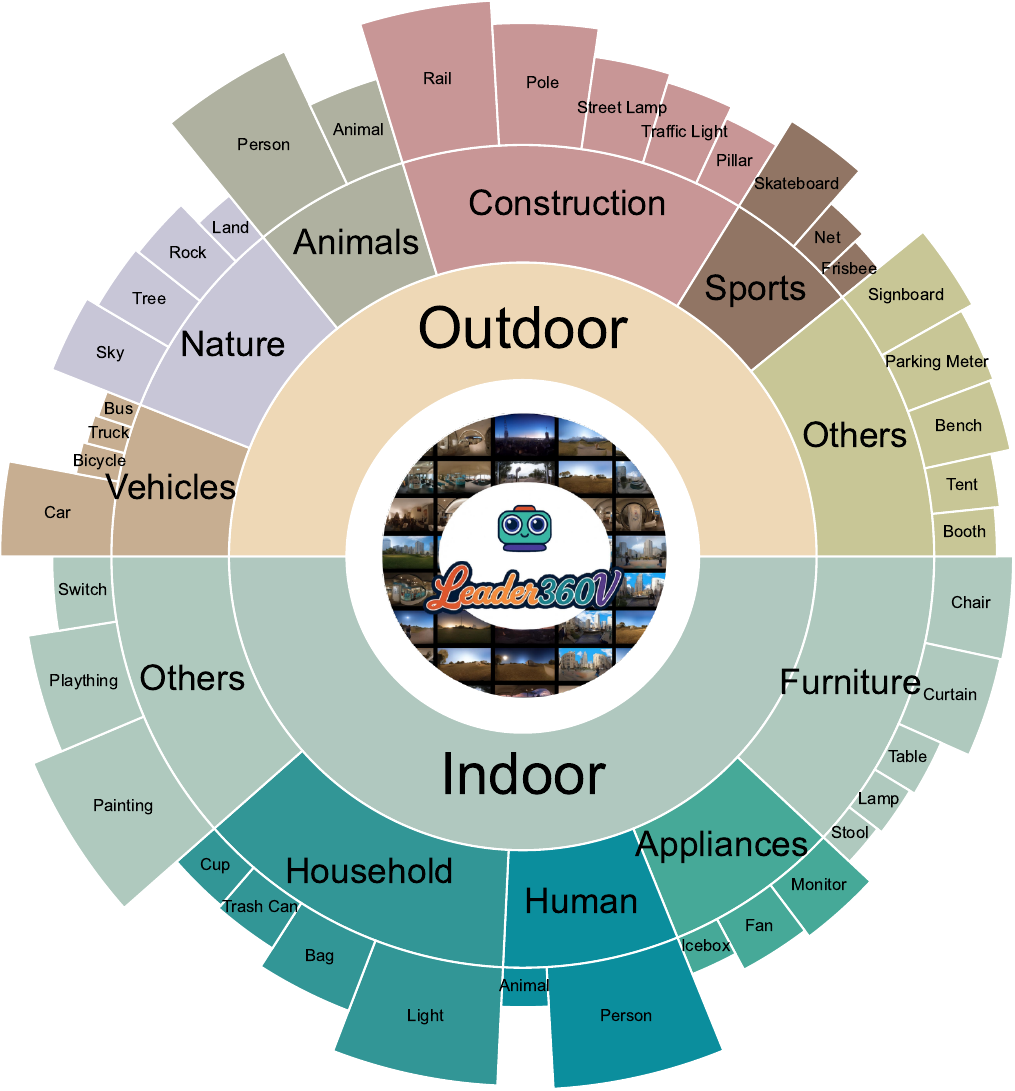}
  \end{center}
  \vspace{-12pt}
  \caption{\small{Category distribution of Leader360V dataset.}}
\vspace{-15pt}
\label{fig:Category}
\end{wrapfigure}

\textbf{Higher Diversity in Scenarios, Especially Cityscape.} Existing 360 video datasets provide limited coverage of cityscape scenarios. This gap hinders the development of practical 360VOTS applications in real-world urban environments. Therefore, we prioritize the inclusion of a wide range of urban environments in Leader360V, capturing variations in architectural styles, traffic conditions, and other dynamic elements. Our Leader360V is also rich in categories, as shown in Fig.~\ref{fig:Category}

\section{Methodology}
\label{sec:Model_Design}

\subsection{The Leader360V Dataset}
\label{sec: Leader360V dataset}

To address the scarcity of large-scale 360 video datasets, we present \textbf{Leader360V}—the first real-world dataset of this scale with diverse scene dynamics and comprehensive annotations for instance segmentation (Leader360V-S) and tracking (Leader360V-T).

\subsubsection{Date Source Analysis}
The Leader360V dataset includes videos collected from existing 360 video datasets, such as 360VOTS~\cite{Xu2024360VOTSVO}, PanoVOS~\cite{Yan2023PanoVOSBN}, \etc. The specific information is shown in Tab. ~\ref{table:datasource}. To address the limited scene diversity in prior datasets, we additionally collect new videos and relabel existing videos. Departing from the collection protocols used in previous 360 video datasets, our self-collected videos exhibit three distinctive properties, as described below.

\textbf{Richer Data Acquisition Methods.}
We employ a variety of recording techniques to capture diverse camera motion patterns, including static camera setups, handheld recordings by moving photographer, and vehicle-based capture. In contrast to previous datasets that rely on limited recording methods, our approach enriches the diversity of 360 videos by simulating a wider range of real-world camera movements, as shown in Tab. \ref {table:comparison_360video}.

\textbf{More Various Perspectives.}
We collect data from multiple viewpoints and perspectives within each scenario. For example, in vehicle-based videos, we include footage from both the roof and the side of the car. This multi-perspective collection is often ignored by previous works.

\subsubsection{Pre-Processing}
% \label{sec:datastatics}
All videos in our Leader360V, whether sourced from existing datasets or self-collected, underwent a standardized pre-processing stage to ensure consistency and quality within the Leader360V dataset. This process included video resizing ($2048\times 1024$), video clipping, face anonymization, and other privacy-preserving operations. Additionally, we removed biased videos and balanced the distribution of different scenarios.
Videos shorter than 5 seconds were excluded, and videos exceeding 30 seconds were clipped to a maximum duration of 30 seconds. The resulting videos range from 10 to 20 seconds in length, with an average duration of 15 seconds. To protect the privacy of both camera operators and passersby, we anonymized the videos by detecting faces in each frame and applying blurring filters, following a procedure similar to that described in~\cite{chen2024360+}.
% Videos shorter than 10 seconds were excluded and videos exceeding 20 seconds were clipped to a maximum duration of 20 seconds. The resulting videos range from 10 to 20 seconds in length, with an average duration of 15 seconds. To protect the privacy of pedestrians, we implemented face censoring to obscure identifiable features. In addition to these steps, we adhered to two primary video selection principles (detailed in the Appendices):

\begin{figure}[t!]
    \centering
\includegraphics[width=\linewidth]{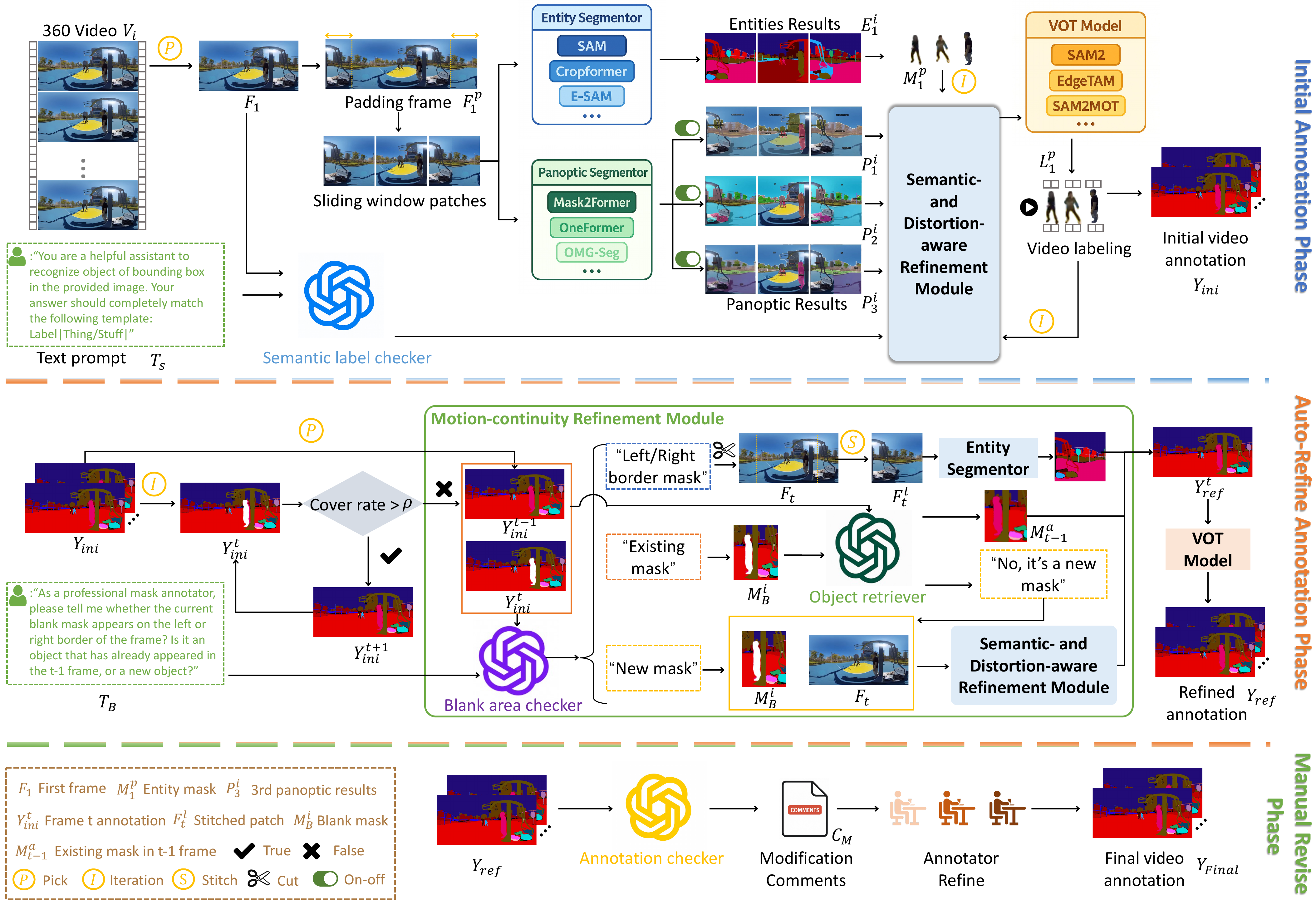}
% \vspace{-15pt}
\caption{\small{The overall of our A$^3$360V pipeline. It consists of three phases: Initial Annotation Phase, Auto-Refine Annotation Phase, and Manual Revise Phase.}}
\vspace{-10pt}
\label{fig: automatic system}
\end{figure}

% \textbf{Video of Objects with Obvious Movement. } We prioritized videos exhibiting significant camera or object motion to facilitate the video quality especially for VOT tasks. Videos such as still natural scenes is excluded due to this rule.

% \textbf{Videos without Object-Intensive Scenarios. } We excluded videos with excessively dense object arrangements to reduce annotation complexity and focus on scenarios with clear object boundaries and manageable occlusion levels.

\subsection{Automatic Annotation Pipeline}
\label{sec: Automatic Annotation Pipline}

Due to the inherently large field of view (FoV), severe geometric distortion, and content discontinuities, annotating 360 video becomes particularly challenging and labor-intensive.
To alleviate the burden on human annotators, we propose the Automatic Annotate Any 360 Video (A$^3$360V) pipeline, as shown in Fig.~\ref{fig: automatic system}, which efficiently integrates pre-trained 2D segmentors and large language models (LLMs) to streamline the labeling process. A$^3$360V operates through a three-stage pipeline: Initial Annotation Phase (Sec. \ref{sec: Initial Annotation Phase}), Auto-Refine Annotation Phase (Sec. \ref{sec: Auto-Refine Annotation Phase}), and Manual Revise Phase (Sec. \ref{sec: Manual Revise Phase}), which will be introduced in detail below.

% \begin{figure}[t!]
%     \centering
% \includegraphics[width=\linewidth]{SDR_Module.pdf}
% % \vspace{-15pt}
% \caption{\small{The overall of our A$^3$360V system.}}
% % \vspace{-10pt}
% \label{fig: automatic system}
% \end{figure}

\subsubsection{Initial Annotation Phase}
\label{sec: Initial Annotation Phase}

In the Initial Annotation Phase, given a 360 video \(\mathcal{V}_i\), A$^3$360V begins by selecting the first frame \(F_1\) as the starting point for annotation.
% The segmentation result of \(F_1\) is then used as a prompt to guide a video object tracking (VOT) model, which propagates the annotations to subsequent frames in the sequence.
To mitigate the issue of horizontal content discontinuity caused by ERP—particularly at the left and right image borders, we first apply horizontal padding to \(F_1\), resulting in an extended frame denoted as \(F_1^p\). The \(F_1^p\) is then divided into a series of overlapping patches using a horizontal sliding window.
% This strategy addresses the large field of view (FoV) inherent to ERP by partitioning the frame into spatially localized regions that match the receptive scale of conventional 2D segmentors, thereby improving their ability to detect small or distant objects that might otherwise be overlooked.
Each of these patches is subsequently processed by a diverse set of pre-trained segmentors, which we categorize into two groups: the first group comprises entity segmentors (e.g., SAM \cite{kirillov2023segment}, CropFormer \cite{qi2022high}, E-SAM \cite{zhang2025sam}), which produce class-agnostic instance-level masks capturing perceptual entities without relying on predefined taxonomies. We denote their output on frame \(F_1\) as \(\mathcal{E}_1^i\).
The second group consists of panoptic segmentors (e.g., Mask2Former \cite{cheng2021mask2former, cheng2021maskformer}, OneFormer \cite{jain2023oneformer}, OMG-Seg \cite{li2024omg}), each trained on different datasets to generate class-aware predictions. We denote one model's output on frame \(F_1\) as \(\mathcal{P}_1^i\). These models produce segmentation results aligned with various large label spaces (e.g., COCO \cite{lin2014microsoft}, ADE20K \cite{zhou2019semantic}, and Cityscapes \cite{Cordts2016Cityscapes}), enriching the annotation pool with complementary semantic categories.

% Each of these patches is subsequently processed by a diverse set of pre-trained segmentors, categorized into two groups: \textbf{entity segmentors} (e.g., SAM, CropFormer, E-SAM), which produce class-agnostic instance masks, and \textbf{panoptic segmentors}, which generate class-aware predictions aligned with specific taxonomies.
% The entity segmentors (e.g., SAM, CropFormer, E-SAM) focus on generating instance-level masks that capture perceptual entities without relying on fixed categories. The output is denoted as \(\mathcal{E}_1^i\), representing the raw entity mask predictions for frame \(F_1\).
% In parallel, we employ multiple panoptic segmentors (e.g., Mask2Former, OneFormer, OMG-Seg), each trained on different datasets, to provide diverse class-aware segmentation outputs. Specifically, we denote the outputs as \(\mathcal{CO}_1^i\), \(\mathcal{A}_1^i\), and \(\mathcal{CS}_1^i\), corresponding to predictions aligned with COCO, ADE20K, and Cityscapes taxonomies, respectively.
\begin{wrapfigure}{l}{0.7\textwidth}
\vspace{-25pt}
  \begin{center}
    \includegraphics[width=0.7\textwidth]{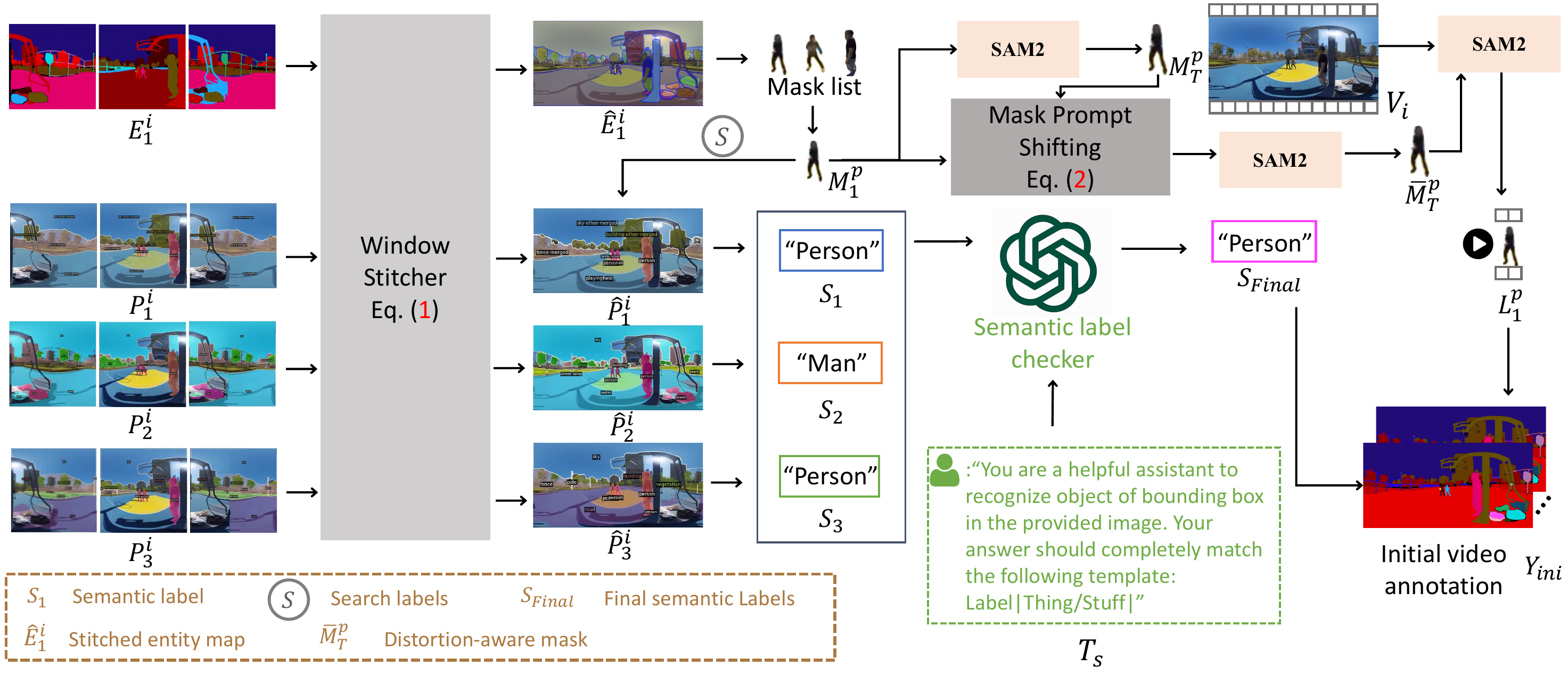}
  \end{center}
  \vspace{-10pt}
  \caption{\small{Illustration of the process of SDR Module.}}
% \vspace{-10pt}
\label{fig: SDR Module}
\vspace{-10pt}
\end{wrapfigure}

To address the object distortion introduced by ERP and unify the semantics across heterogeneous predictions from multiple segmentors, we propose the \textbf{Semantic- and Distortion-aware Refinement (SDR) Module}, as illustrated in Fig.~\ref{fig: SDR Module}.
This module plays a central role in the Initial Annotation Phase by consolidating outputs from both entity and panoptic segmentors into a coherent, distortion-aware annotation for the first frame \(F_1\). While the framework supports an arbitrary number of panoptic segmentors, we illustrate our approach using three representative models in this paper.
Based on the overlapping patch divisions, the SDR module first aggregates patch-wise predictions from the 2D segmentors into full-frame segmentation maps, denoted as \(\mathcal{E}_1^i\), \(\mathcal{P}_1^i\), \(\mathcal{P}_2^i\), and \(\mathcal{P}_3^i\), using a window-stitching operation defined in Eq.~(1).
\begin{equation}
\small
\text{Match}(M_k, M_l) =
\begin{cases}
1, & \text{if } \text{IoU}(M_k, M_l) > \tau, \\
0, & \text{otherwise}
\end{cases} \tag{1}
\end{equation}
where \(M_k\) and \(M_l\) denote instance masks predicted from overlapping regions of different patches, and \(\tau\) is a predefined threshold to determine whether two masks represent the same object.

To resolve class labeling inconsistency, we incorporate a large language model (LLM)-based semantic label checker within SDR. For each entity mask proposal from \(\mathcal{E}_1^i\), the pipeline retrieves corresponding label candidates from \(\mathcal{P}_1^i\), \(\mathcal{P}_2^i\), and \(\mathcal{P}_3^i\), and feeds them into the semantic label checker via a structured prompt \(T_s\). The semantic label checker selects the most semantically appropriate label, yielding a harmonized set of final semantic labels for all entities.
To obtain distortion-aware masks, we leverage the robustness of video foundation models by feeding each \(M_1^p\) as a mask prompt into the model in an iterative manner. Taking SAM2~\cite{Ravi2024SAM2S} as an example, we first input \(M_1^p\) to obtain a coarse prediction \(M_T^p\). To improve its reliability under 360 distortion, we perform \textit{Mask Prompt Shifting} by applying spatial shifts to \(M_T^p\) and refeeding the shifted masks into SAM2. Since SAM2 returns a single mask per query, this process yields a set of candidate masks \(\mathcal{M}_T^p = \{ M_T^{p, \delta} \mid \delta \in \mathcal{D} \}\). We then select the most frequently returned result as the final refined mask \(\bar{M}_T^p\):
\begin{equation}
\bar{M}_T^p = \mathop{\arg\max}_{M \in \mathcal{M}_T^p} \; \sum_{\delta \in \mathcal{D}} \mathbb{I} \left[ \text{IoU}\left(M, \text{SAM2}(\text{Shift}(M_1^p, \delta)) \right) > \tau \right] \tag{2},
\label{equ: 1}
\end{equation}
\(\mathcal{D}\) denotes the set of shift directions, \(\mathbb{I}[\cdot]\) is the indicator function, and \(\tau\) is the IoU threshold for mask consistency.
The \(\bar{M}_T^p\) is used to track the entity across subsequent frames, generating a sequence of annotations \(\mathcal{L}_1^p\), which, combined with LLM-verified labels, form the initial video annotation \(Y_{\text{ini}}\).

\subsubsection{Auto-Refine Annotation Phase}
\label{sec: Auto-Refine Annotation Phase}

Given the initial annotated frame \( Y_{\text{ini}} \) produced in the previous stage, the Auto-Refine Annotation Phase aims to propagate and correct annotations across the remaining frames in the 360 video \(\mathcal{V}_i\). This stage iteratively processes each frame \( F_t \) using a coverage-guided strategy and performs dynamic refinement for missing or misaligned regions.
At each timestamp \(t\), we evaluate the coverage rate of the current annotation \(Y^t_{\text{ini}}\) against a predefined threshold \(\rho\). If the coverage is sufficient (i.e., coverage rate \(> \rho\)), we accept \(Y^t_{\text{ini}}\) and use it to generate the initial annotation for the next frame, \(Y^{t+1}_{\text{ini}}\). Otherwise, the \textbf{Motion-Continuity Refinement (MCR) Module} is triggered to improve the annotation quality before propagation.

To identify unannotated areas in the current frame, we employ an LLM-based agent, referred to as the \textbf{Blank Area Checker}, which is guided by a task-specific text prompt \(T_B\). The prompt instructs the agent to infer the nature of each blank region. Based on this semantic inquiry, the blank region is classified into one of three types:
\textbf{1. Left/Right Border Mask:} If the blank region lies near the left or right boundary of \(Y^t_{\text{ini}}\), we crop and horizontally stitch the current frame \(F_t\) to form a complete view of the context \(F_t^l\). This operation addresses the content discontinuities inherent in ERP, allowing entity segmentors to reprocess the region with improved spatial continuity.
\textbf{2. Existing Mask:} If the blank mask \(M_{B}^i\) corresponds to a previously annotated object \(M_{t-1}^a\), we invoke an LLM-based agent, referred to as the \textbf{Object Retriever}, to search for a matching mask within the prior frame's annotation \(Y^{t-1}_{\text{ini}}\). If a match is successfully retrieved, the blank region inherits the same semantic label. Otherwise, it is reclassified as a new mask.
\textbf{3. New Mask:} If the area represents a newly emerged object not seen in earlier frames, we treat the \(M_{B}^i\) as a novel instance and re-enter it, together with the current frame \(F_t\), into the SDR module. This process yields a refined entity segmentation and assigns an accurate semantic label.
After resolving all incomplete regions, the annotated frame is updated to \( Y^t_{\text{ref}} \). This refined annotation is then passed to a \textbf{VOT Model} (e.g., SAM2 \cite{Ravi2024SAM2S}, EdgeTAM \cite{zhou2025edgetam}, SAM2MOT\cite{jiang2025sam2mot}) for temporal smoothing and consistency adjustment. The final result is appended to the refined annotation set \( Y_{\text{ref}} \), which accumulates high-quality annotations.

\subsubsection{Manual Revise Phase}
\label{sec: Manual Revise Phase}
\begin{figure}[t!]
    \centering
    % \vspace{-25pt}
\includegraphics[width=\linewidth]{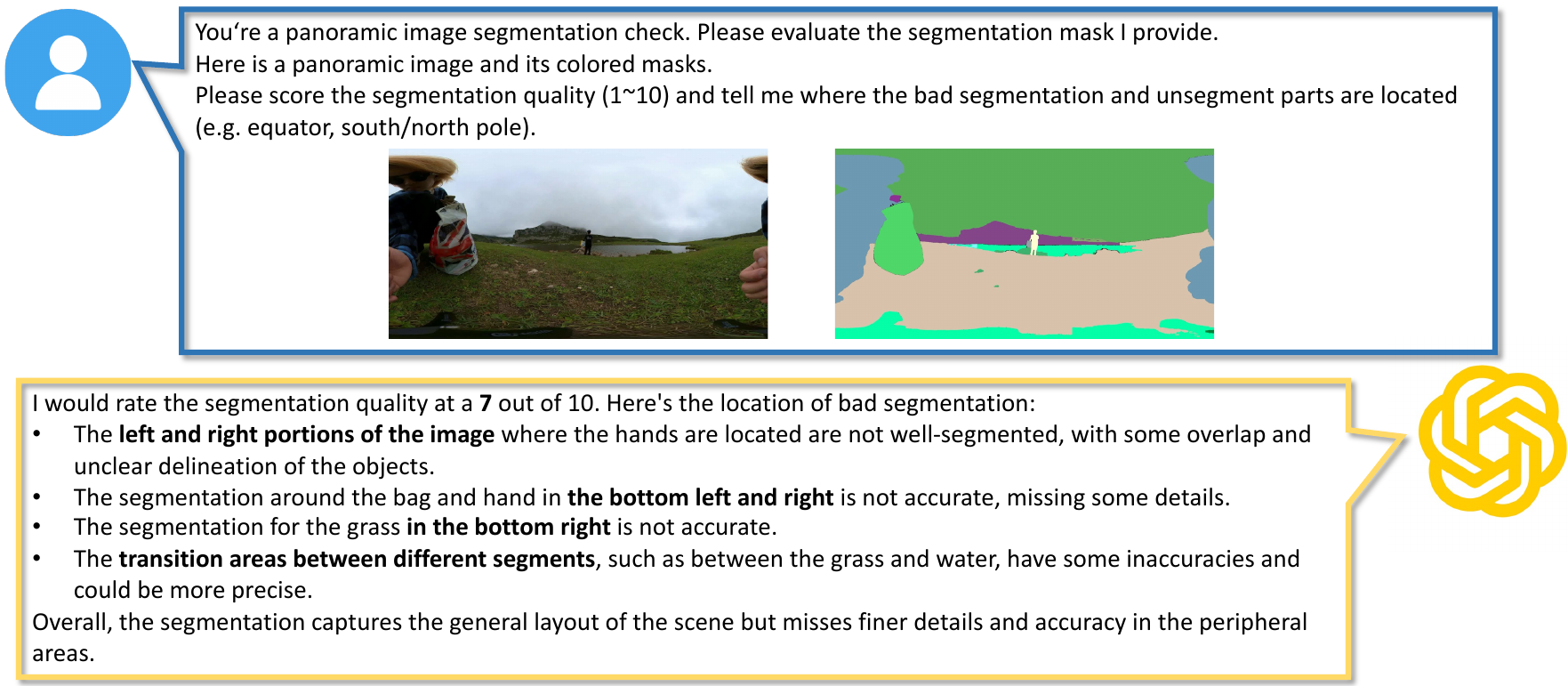}
\caption{\small{Visualization of an example of the feedback provided by the annotation checker, who scores the annotation and points out where bad annotations lie.}}
% \vspace{-10pt}
\label{fig: annotation_checker}
\end{figure}

Although the Auto-Refine phase significantly reduces the need for human intervention, ensuring high-quality and consistent annotations across the entire video \(\mathcal{V}_i\) still requires a final verification step. In this stage, we introduce an LLM-based agent, referred to as \textbf{Annotation Checker}, which analyzes the refined annotation \(Y_{\text{ref}}\) and generates natural language modification suggestions, denoted as \(C_M\). These comments highlight potential issues in spatial consistency, class accuracy, or temporal coherence, as shown in Fig.~\ref{fig: annotation_checker}.
A group of human annotators then reviews and edits \(Y_{\text{ref}}\) based on the LLM-generated feedback \(C_M\), making targeted refinements rather than re-annotating from scratch. This human-in-the-loop revision process results in the final high-fidelity annotation set, denoted as \(Y_{\text{final}}\).

\section{Experiment}
\label{experiment}

\subsection{Implementation details}
\noindent\textbf{Auto-Annotation Settings.}
During the dataset construction, we employ CropFormer \cite{qi2022high} as the entity segmentation model and OneFormer \cite{jain2023oneformer} as the panoptic segmentation model. % Specifically, for CropFormer, we utilize the variant based on the HorNet-L backbone \cite{rao2022hornet}. For OneFormer, we select the models demonstrating optimal performance on the COCO \cite{lin2014microsoft}, ADE20K \cite{zhou2019semantic}, and Cityscapes \cite{Cordts2016Cityscapes}.
Furthermore, GPT-4o \cite{achiam2023gpt} is incorporated as an LLM to function as a checker for semantic labels, blank areas, and annotations.

\begin{wraptable}{l}{0.4\linewidth}
\vspace{-15pt}
\centering
\caption{\small{Evaluation on samples of Leader360V for SAM \cite{kirillov2023segment} -based methods.}}
\tiny
\begin{tabular}{cc|ccc}
\toprule
\multicolumn{2}{c|}{\multirow{3}{*}{\textbf{Model}}} & \multicolumn{3}{c}{\textbf{Leader360V-S Test}} \\
\cdashline{3-5}
\multicolumn{2}{c|}{} & \multirow{2}{*}{$\mathcal{J} \& \mathcal{F}\uparrow$}&
\multirow{2}{*}{$\mathcal{J}\uparrow$}         & \multirow{2}{*}{$\mathcal{F}\uparrow$}          \\
\multicolumn{2}{l|}{}                                 &                      &                    &            \\
\midrule
\midrule
\multicolumn{2}{c|}{\textbf{PerSAM\cite{zhang2023personalize}}}&18.7 &13.1
&24.3\\
\multicolumn{2}{c|}{\textbf{SAM-PT\cite{sam-pt}}}&45.1 &37.8
&52.4\\
\multicolumn{2}{c|}{\textbf{GoodSAM\cite{zhang2024goodsam}}}& 27.4 & 20.9       & 33.9  \\
\midrule
\end{tabular}
\label{table: sam_based_result}
\vspace{-10pt}
\end{wraptable}

\noindent\textbf{Evaluation Subset.}
We selected 500 videos as our sample dataset, ensuring that the distribution of scenarios and categories was similar to that of the entire dataset. Inspired by 360VOTS \cite{Xu2024360VOTSVO} and PanoVOS* \cite{Yan2023PanoVOSBN}, we divided the 500 videos into a training set (250), a validation set (125), and a test set (125). For the validation set and test set, 66\% of the clips are clipped from the original train set videos as val and test sets, and the rest are used as the train set. The remaining clips in the validation and test sets were selected from new and unseen scenarios.

\noindent\textbf{Evaluation Metric.}
For VOS task, we choose region accuracy ($\mathcal{J}$), boundary accuracy ($\mathcal{F}$), and combined average ($\mathcal{J} \& \mathcal{F}$) as evaluation metrics, following the standard protocol \cite{perazzi2016benchmark, Yan2023PanoVOSBN}. For the VOT task, we utilize metrics of dual success ($S_{\textit{dual}}$) and dual precision ($P_{\textit{dual}}$), following 360VOTS \cite{Xu2024360VOTSVO}.

\subsection{Comparison Result Analysis}

\noindent\textbf{Results via SAM-based Model.}
Inspired by \cite{Yan2023PanoVOSBN}, we assess various SAM \cite{kirillov2023segment} versions on our Leader-360V test set, as shown in Tab. \ref{table: sam_based_result}. Due to the domain gap between 2D and 360 images, PerSAM \cite{zhang2023personalize} shows poor performance. Similarly, SAM-PT \cite{sam-pt}, a SAM-based VOS model, also delivers unsatisfactory results. Additionally, GoodSAM \cite{zhang2024goodsam}, a 360 image segmentation model, is evaluated and yields disappointing outcomes. These results highlight the need for further exploration to bridge the domain gap and improve tracking performance for 360 videos.

\noindent\textbf{Results of VOS Task.}
We demonstrate the effectiveness of the Leader360V dataset for the VOS task in Tab. \ref{table: VOS_comparision}. While traditional 2D models show unsatisfactory performance on 360 video (e.g., XMem at \textbf{42.4} in terms of $\mathcal{J} \& \mathcal{F}$), PSCFormer, trained specifically on our train subset, exhibits significant improvement (\textbf{+36.3} for $\mathcal{J} \& \mathcal{F}$). This highlights the necessity of Leader360V for the 360VOS task.

\begin{table}[h]
\begin{minipage}{0.5\linewidth}
\centering
\caption{\small{Qualitative comparison between VOS models for 2D video and 360 video on samples of the Leader360V dataset.}}
\tiny
\setlength{\tabcolsep}{0.53mm}
\begin{tabular}{c|cc|ccc}
\toprule
\multirow{3}{*}{\textbf{Task}} & \multicolumn{2}{c|}{\multirow{3}{*}{\textbf{Model}}} & \multicolumn{3}{c}{\textbf{Leader360V-S Test}} \\
\cdashline{4-6}
& \multicolumn{2}{c|}{} & \multirow{2}{*}{$\mathcal{J} \& \mathcal{F}\uparrow$}&
\multirow{2}{*}{$\mathcal{J}\uparrow$}         & \multirow{2}{*}{$\mathcal{F}\uparrow$}          \\
& \multicolumn{2}{c|}{}                                 &                      &                    &            \\
\midrule
\midrule
\multirow{5}{*}{\textbf{2D}} & \multicolumn{2}{c|}{\textbf{XMem}\cite{cheng2022xmem}}& 42.4       & 35.9  & 48.9\\
& \multicolumn{2}{c|}{\textbf{AOTL}\cite{yang2021associating}} & 43.1       & 37.7  & 48.5\\
& \multicolumn{2}{c|}{\textbf{R50-AOT-L}\cite{yang2021associating}}& 43.9       & 39.0  & 48.8\\
& {\textbf{SwinB-AOT-L (Untrained)} \cite{yang2021associating}}  & & 38.8       & 34.2  & 43.4\\
\rowcolor{tablehighlightgray}
\cellcolor{white} & \cellcolor{white} \textcolor{orange}{\ding{80}}{\textbf{SwinB-AOT-L (Trained)} \cite{yang2021associating}} &\cellcolor{white} & 58.3$\uparrow$\textcolor{blue}{19.5}       & 49.4$\uparrow$\textcolor{blue}{15.2}  & 67.2$\uparrow$\textcolor{blue}{23.8} \\
\midrule
\midrule
\multirow{2}{*}{\textbf{360}} & {\textbf{PSCFormer (Untrained)}} \cite{Yan2023PanoVOSBN}&  & 24.3       & 17.5  & 31.1\\
% \rowcolor{tablehighlightgray}
& \textcolor{orange}{\ding{80}}{\textbf{PSCFormer (Trained)}} \cite{Yan2023PanoVOSBN}& \cellcolor{white}  & \cellcolor{tablehighlightgray}60.6$\uparrow$\textcolor{blue}{36.3}        & \cellcolor{tablehighlightgray}51.5$\uparrow$\textcolor{blue}{34.0}   & \cellcolor{tablehighlightgray}69.7$\uparrow$\textcolor{blue}{38.6} \\
\midrule
\end{tabular}
\label{table: VOS_comparision}
\end{minipage}
\begin{minipage}{0.47\linewidth}
\vspace{-5pt}
\centering
\caption{\small{Qualitative comparison between VOT models for 2D video and 360 video on samples of the Leader360V dataset.}}
\label{table:comparison_VOT}
\tiny
\setlength{\tabcolsep}{0.54mm}
\begin{tabular}{c|cc|c@{\hspace{0.2cm}}c}
\toprule
\multirow{3}{*}{\textbf{Task}} & \multicolumn{2}{c|}{\multirow{3}{*}{\textbf{Model}}} & \multicolumn{2}{c}{\textbf{Leader360V-S Test}} \\
\cdashline{4-5}
& \multicolumn{2}{c|}{} & \hspace{0.1cm} \multirow{2}{*}{$S_{\textit{dual}}\uparrow$}&
\multirow{2}{*}{$P_{\textit{dual}}\uparrow$}         \\
& \multicolumn{2}{l|}{}                                 &                      & \\
\midrule
\midrule
\multirow{5}{*}{\textbf{2D}} & \multicolumn{2}{c|}{\textbf{SiamX}\cite{9812327}} &  \hspace{0.1cm}0.217       & 0.183\\
& \multicolumn{2}{c|}{\textbf{AiATrack}\cite{gao2022aiatrack}}&  \hspace{0.1cm}0.286       & 0.252\\
& \multicolumn{2}{c|}{\textbf{ProContEXT}\cite{lan2023procontext}}&  \hspace{0.1cm}0.308       & 0.270\\
& {\textbf{SimTrack (Untrained)}\cite{chen2022backbone}} &  &  \hspace{0.1cm}0.291       & 0.256\\
\rowcolor{tablehighlightgray}
\cellcolor{white} & \cellcolor{white} \textcolor{orange}{\ding{80}}{\textbf{SimTrack (Trained)}\cite{chen2022backbone}}& \cellcolor{white} &  \hspace{0.1cm}0.417$\uparrow$\textcolor{blue}{12.6}       & 0.373$\uparrow$\textcolor{blue}{11.7}\\
\midrule
\midrule
\textbf{360} & \multicolumn{2}{c|}{\textbf{SiamX-360}\cite{hhuang2022siamx}}&  \hspace{0.1cm}0.282       & 0.254\\
\midrule
\end{tabular}
\label{table: VOT_comparision}
\end{minipage}
\vspace{-5pt}
\end{table}

\noindent\textbf{Results of VOT Task.}
Tab. \ref{table:comparison_VOT} presents comparison results among several trackers for both 2D and 360 tasks. Based on the quantitative results, it is evident that our dataset significantly enhances the performance of the tracker. The performance of 2D tracker SimTrack \cite{chen2022backbone}, which is originally designed for 2D video tasks, is obviously improved, \textbf{+12.6} for $S_{\textit{dual}}$ and \textbf{+11.7} for $P_{\textit{dual}}$. However, the performance of the popular 360 model \cite{hhuang2022siamx} on our dataset does not meet our expectations.

\begin{table}[]
\centering
\caption{Evaluation of domain transfer (from YouTubeVOS \cite{xu2018youtube} to our Leader360V Test).}
\small
\setlength{\tabcolsep}{0.6mm}
\begin{tabular}{c|cc|ccc|ccc}
\hline
\multirow{3}{*}{Methods}  & \multicolumn{2}{c|}{Training Dataset} & \multicolumn{3}{c|}{YouTubeVOS Test} & \multicolumn{3}{c}{Leader360V Test} \\
\cdashline{2-9}
                        & \multirow{2}{*}{YouTubeVOS}      & \multirow{2}{*}{Self-Collected}      & \multirow{2}{*}{$\mathcal{J}\&\mathcal{F}\uparrow$}          & \multirow{2}{*}{$\mathcal{J}\uparrow$}         & \multirow{2}{*}{$\mathcal{F}\uparrow$}         & \multirow{2}{*}{$\mathcal{J}\&\mathcal{F}\uparrow$}          & \multirow{2}{*}{$\mathcal{J}\uparrow$}         & \multirow{2}{*}{$\mathcal{F}\uparrow$}\\
& & & & & & & & \\\midrule \midrule
\multirow{2}{*}{\textbf{AOT}\cite{yang2021associating}}    &\ding{51}                 &\ding{55}                     & 72.8             &  64.2         &  84.1         &  42.9           & 37.5          & 48.3          \\
                        & \ding{51}                &\ding{51}                     & 61.4 $\downarrow$\textcolor{red}{11.4}            & 53.7$\downarrow$\textcolor{red}{10.5}          & 69.1$\downarrow$\textcolor{red}{15.0}          & 51.6$\uparrow$\textcolor{blue}{8.7}           &  44.7$\uparrow$\textcolor{blue}{7.2}          & 58.5$\uparrow$\textcolor{blue}{10.2}           \\ \midrule
\multirow{2}{*}{\textbf{STCN} \cite{cheng2021stcn}}   &\ding{51}                 &\ding{55}                     &75.8              & 68.9          &82.9           &47.2             &38.8           &55.6           \\
                        &\ding{51}                &\ding{51}                     &60.1$\downarrow$\textcolor{red}{15.7}              & 51.6$\downarrow$\textcolor{red}{17.3}          &68.6$\downarrow$\textcolor{red}{14.3}           &55.4$\uparrow$\textcolor{blue}{8.2} &49.8$\uparrow$\textcolor{blue}{11.0}           &61.0$\uparrow$\textcolor{blue}{5.4}           \\ \midrule
\multirow{2}{*}{\textbf{RDE}\cite{li2022recurrent}}    &\ding{51}                 &\ding{55}                     &61.3              & 54.9          & 67.7          &36.2             &30.1           & 42.3          \\
                        &\ding{51}                &\ding{51}                     & 48.9$\downarrow$\textcolor{red}{12.4}             & 41.8$\downarrow$\textcolor{red}{13.1}          & 56.0$\downarrow$\textcolor{red}{11.7}          & 45.5$\uparrow$\textcolor{blue}{9.3}            & 37.6$\uparrow$\textcolor{blue}{7.5}           & 53.4$\uparrow$\textcolor{blue}{11.1}          \\ \midrule
\multirow{2}{*}{\textbf{XMem}\cite{cheng2022xmem}}   &\ding{51}                 &\ding{55}                     & 76.6             & 73.3          & 79.9          & 42.5            & 35.8          & 49.2          \\
                        &\ding{51}                &\ding{51}                     & 62.5$\downarrow$\textcolor{red}{14.1}              & 57.4$\downarrow$\textcolor{red}{15.9}          & 67.6$\downarrow$\textcolor{red}{12.3}          &  55.1$\uparrow$\textcolor{blue}{12.6}           & 49.7$\uparrow$\textcolor{blue}{13.9}          & 60.5$\uparrow$\textcolor{blue}{11.3}          \\ \midrule
\multirow{2}{*}{\textbf{XMem++}\cite{bekuzarov2023xmem++}} &\ding{51}                 &\ding{55}                     & 77.5             & 74.2          & 80.8          & 45.9            & 39.0          & 52.8          \\
                        &\ding{51}              &\ding{51}                    & 62.7$\downarrow$\textcolor{red}{14.8}              & 58.6$\downarrow$\textcolor{red}{15.6}          & 66.8$\downarrow$\textcolor{red}{14.0}          & 56.3$\uparrow$\textcolor{blue}{10.4}            & 50.6$\uparrow$\textcolor{blue}{11.6}          & 62.0$\uparrow$\textcolor{blue}{9.2}          \\ \midrule
\end{tabular}
\label{table: VOS_transfer}
\end{table}

\begin{table}[]
\centering
\caption{Evaluation of domain transfer (from TrackingNet \cite{muller2018trackingnet} to our Leader360V Test).}
\setlength{\tabcolsep}{0.6mm}
\begin{tabular}{c|cc|cc|cc}
\hline
\multirow{3}{*}{Methods}  & \multicolumn{2}{c|}{Training Dataset} & \multicolumn{2}{c|}{TrackingNet  Test} & \multicolumn{2}{c}{Leader360V Test} \\
\cdashline{2-7}
                        & \multirow{2}{*}{TrackingNet }      & \multirow{2}{*}{Self-Collected}      & \multirow{2}{*}{$S_{\textit{dual}}\uparrow$}          & \multirow{2}{*}{$P_{\textit{dual}}\uparrow$}         & \multirow{2}{*}{$S_{\textit{dual}}\uparrow$}          & \multirow{2}{*}{$P_{\textit{dual}}\uparrow$}\\
& & & & & & \\\midrule \midrule
\multirow{2}{*}{\textbf{SiamX}\cite{9812327}}    &\ding{51}                 &\ding{55}                     & 75.7             &  72.2         & 20.6          & 17.0                      \\
                        & \ding{51}                &\ding{51}                     & 60.8$\downarrow$\textcolor{red}{14.9}             & 56.4$\downarrow$\textcolor{red}{15.8}          & 30.9$\uparrow$\textcolor{blue}{10.3}          & 28.3$\uparrow$\textcolor{blue}{11.3}          \\ \midrule
\multirow{2}{*}{\textbf{AiATrack}\cite{gao2022aiatrack}}   &\ding{51}                 &\ding{55}                     & 82.3             & 79.8          &  28.0         &  24.7                      \\
                        &\ding{51}                &\ding{51}                     &  64.5$\downarrow$\textcolor{red}{17.8}            &  59.8$\downarrow$\textcolor{red}{20.0}         & 36.1$\uparrow$\textcolor{blue}{12.1}          &  33.6$\uparrow$\textcolor{blue}{8.9}                      \\ \midrule
\multirow{2}{*}{\textbf{ProContEXT}\cite{lan2023procontext}}    &\ding{51}                 &\ding{55}                     &  84.2            & 81.8          & 30.5          & 26.9                       \\
                        &\ding{51}                &\ding{51}                     & 64.7$\downarrow$\textcolor{red}{19.5}             & 59.1$\downarrow$\textcolor{red}{22.7}          & 42.4$\uparrow$\textcolor{blue}{11.9}          & 37.9$\uparrow$\textcolor{blue}{11.0}                      \\ \midrule
\multirow{2}{*}{\textbf{MixFormer}\cite{cui2022mixformer}}   &\ding{51}                 &\ding{55}                     & 83.7             & 80.9          & 29.6         & 24.4                       \\
                        &\ding{51}                &\ding{51}                     & 65.0$\downarrow$\textcolor{red}{18.7}             & 60.2$\downarrow$\textcolor{red}{20.7}          & 41.7$\uparrow$\textcolor{blue}{12.1}          & 38.3$\uparrow$\textcolor{blue}{13.9}                      \\ \midrule
\multirow{2}{*}{\textbf{SimTrack}\cite{chen2022backbone}} &\ding{51}                 &\ding{55}                     & 83.1             & 80.6          & 29.1          &  24.0                     \\
                        &\ding{51}                &\ding{51}                    & 63.3$\downarrow$\textcolor{red}{19.8}             & 58.8\textcolor{red}{21.8}          & 41.5$\uparrow$\textcolor{blue}{12.4}          &  37.6$\uparrow$\textcolor{blue}{13.6}                    \\ \midrule
\end{tabular}
\label{table: VOT_transfer}
\end{table}
\noindent\textbf{Results of domain transfer evaluations.}
Tab.~\ref{table: VOS_transfer} illustrates the domain transfer results of state-of-the-art video object segmentation (VOS) models from conventional planar-domain datasets (e.g., YouTubeVOS\cite{xu2018youtube}) to our panoramic Leader360V benchmark. We observe a significant performance degradation across all methods when directly applying models trained on YouTubeVOS to 360 video content. Specifically, the combined region and boundary accuracy metric $\mathcal{J}\&\mathcal{F}$ consistently drops by over 10–18 points. This highlights the substantial domain gap between conventional narrow field-of-view videos and equirectangular panoramic inputs, which introduce geometric distortion, boundary discontinuities, and changes in object appearance and motion patterns.
However, after fine-tuning these models on the training split of Leader360V, we observe a notable recovery in performance. The $\mathcal{J}\&\mathcal{F}$ scores increase by 10–14 points across models, demonstrating that VOS methods can adapt effectively to the challenges of 360 scenes when provided with appropriate training data. Notably, methods like XMem \cite{cheng2022xmem} and XMem++ \cite{bekuzarov2023xmem++} exhibit strong adaptability, suggesting that memory-based and transformer-based architectures may be better suited for handling panoramic temporal dynamics. These results emphasize the importance of domain-specific training and the role of a diverse and semantically rich dataset like Leader360V in bridging the generalization gap. Additionally, the improvement indicates that label distribution mismatch—where many rare or scene-specific categories are underrepresented in planar datasets—is another contributing factor, and Leader360V's curated taxonomy helps alleviate this issue.

Tab.~\ref{table: VOT_transfer} presents the domain adaptation analysis for visual object tracking (VOT) models under similar transfer settings. Interestingly, while VOS models demonstrate reasonable recovery after fine-tuning, VOT models suffer even more pronounced performance degradation when evaluated directly on Leader360V. Performance increases of 8–14 points post-finetuning still fall short of the baseline accuracy on planar video benchmarks. This suggests that object tracking in 360 video is inherently more challenging due to compounded issues such as viewpoint wrap-around, severe distortion at the poles, and the loss of consistent object appearance across time.

These findings underscore one key insight:
% (1) while VOS models benefit substantially from domain-specific training, VOT models face greater difficulties in adapting to panoramic inputs, possibly due to stronger reliance on stable visual features and motion continuity assumptions that break down in ERP projections; (2)
Leader360V serves as not only a benchmark but also an effective training resource that improves model generalization and robustness under 360 conditions. Together, these results validate our motivation to construct Leader360V and demonstrate its value in promoting research into panoramic video understanding tasks.

\subsection{Ablation Study}
\begin{figure}[t!]
    \centering
\includegraphics[width=\linewidth]{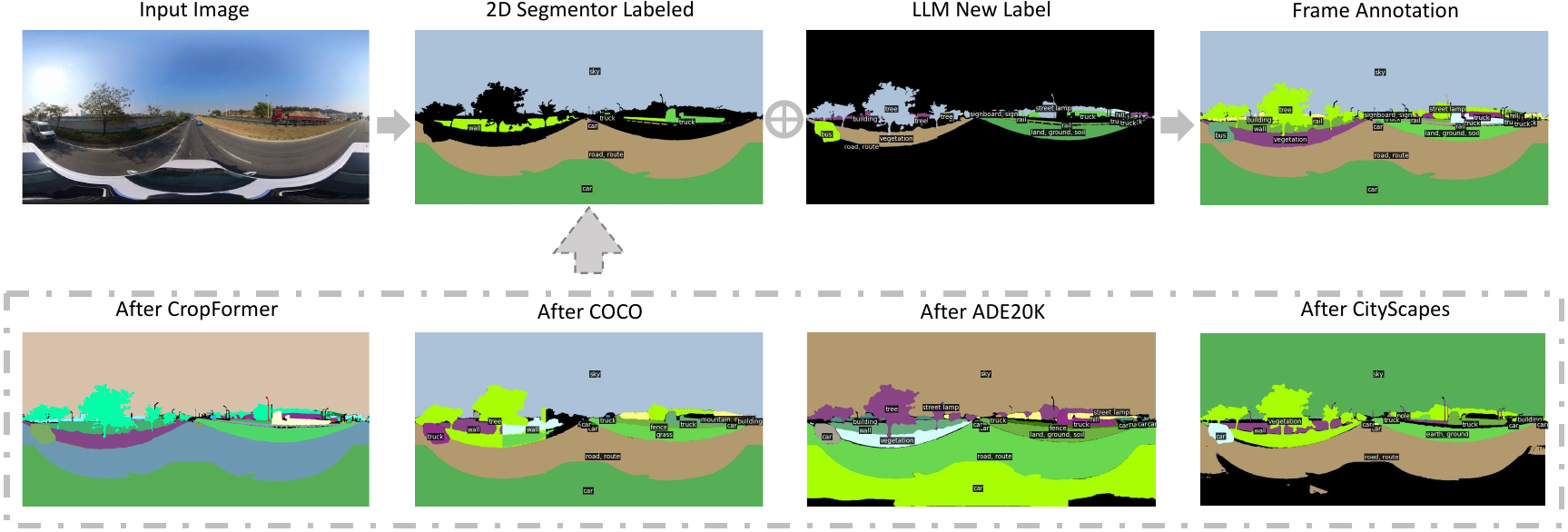}
% \vspace{-15pt}
\caption{\small{Example visualizations of the sequential application of entity segmentor, 2D segmentor, and semantic label checker in SDR Module.}}
\vspace{-15pt}
\label{fig: ablation}
\end{figure}

\noindent\textbf{Effectiveness of Phase I.}
In Tab. \ref{table: component_analysis}, we sampled 100 frames from the dataset that necessitate auto-refinement, using the final annotations as the ground truth benchmark.
\begin{wraptable}{l}{0.45\textwidth}
% \vspace{-10pt}
\begin{minipage}{0.45\textwidth}
\centering
\caption{\small{Comparison of results from different components.}}
\vspace{-5pt}
\tiny
\setlength{\tabcolsep}{0.6mm}
\begin{tabular}{cc|ccc}
\toprule
\multicolumn{2}{c|}{\multirow{3}{*}{\textbf{Phase Component}}} & \multicolumn{3}{c}{\textbf{Updated Frame}} \\
\cdashline{3-5}
\multicolumn{2}{c|}{} & \multirow{2}{*}{$\mathcal{J} \& \mathcal{F}\uparrow$}&
\multirow{2}{*}{$\mathcal{J}\uparrow$}         & \multirow{2}{*}{$\mathcal{F}\uparrow$}          \\
\multicolumn{2}{l|}{}                                 &                      &                    &            \\
\midrule
\midrule
\cellcolor{white} \multirow{2}{*}{\textbf{Phase I}}& 2D Segmentor& 25.1           & 16.4         & 33.8        \\
% \rowcolor{tablehighlightgray}
& After SDR         & \cellcolor{tablehighlightgray}67.3$\uparrow$\textcolor{blue}{42.2}           & \cellcolor{tablehighlightgray}61.1$\uparrow$\textcolor{blue}{44.7}         & \cellcolor{tablehighlightgray}73.5$\uparrow$\textcolor{blue}{39.7}\\
 \midrule
\multirow{2}{*}{\textbf{Phase II}}       & SAM2                &
35.8       & 26.5         & 45.1\\
% \rowcolor{tablehighlightgray}
 & After MCR        & \cellcolor{tablehighlightgray}\textbf{75.9}$\uparrow$\textcolor{blue}{40.1}           & \cellcolor{tablehighlightgray}\textbf{69.3}$\uparrow$\textcolor{blue}{42.8}         & \cellcolor{tablehighlightgray}\textbf{82.5}$\uparrow$\textcolor{blue}{37.4}\\ \midrule
\end{tabular}
\label{table: component_analysis}
\end{minipage}
\vspace{-10pt}
\end{wraptable}
The outputs from the 2D segmentor and the semantic label checker in Phase I, are evaluated. The initial performance of the 2D segmentor is hindered by the exclusion of masks with uncertain labels, resulting in relatively low accuracy scores. Nevertheless, the SDR Module facilitates the assignment of suitable labels to previously unlabeled masks, which substantially enhances the $\mathcal{J} \& \mathcal{F}$ metric by \textbf{+42.2}. An example of mask results at various stages is depicted in Fig.~\ref{fig: ablation}. Upon comparison, it is evident that the 2D segmenter encounters difficulties in annotating novel objects that fall outside its distribution, and some pre-existing labels exhibit low accuracy, especially those located at a distance. The semantic label checker adeptly addresses these challenges by supplementing new labels and unifying existing labels from our category spaces, thereby enhancing overall accuracy. As demonstrated in Fig.~\ref{fig: ablation}, instances initially labeled ambiguously as "tree" and "vegetation" are ultimately unified as "tree." Additionally, the label "signboard, sign," which was overlooked by the 2D segmenter, is successfully added by the semantic label checker.

\begin{figure}[t!]
\begin{minipage}{0.53\linewidth}
\vspace{-5pt}
  \begin{center}
    \includegraphics[width=\linewidth]{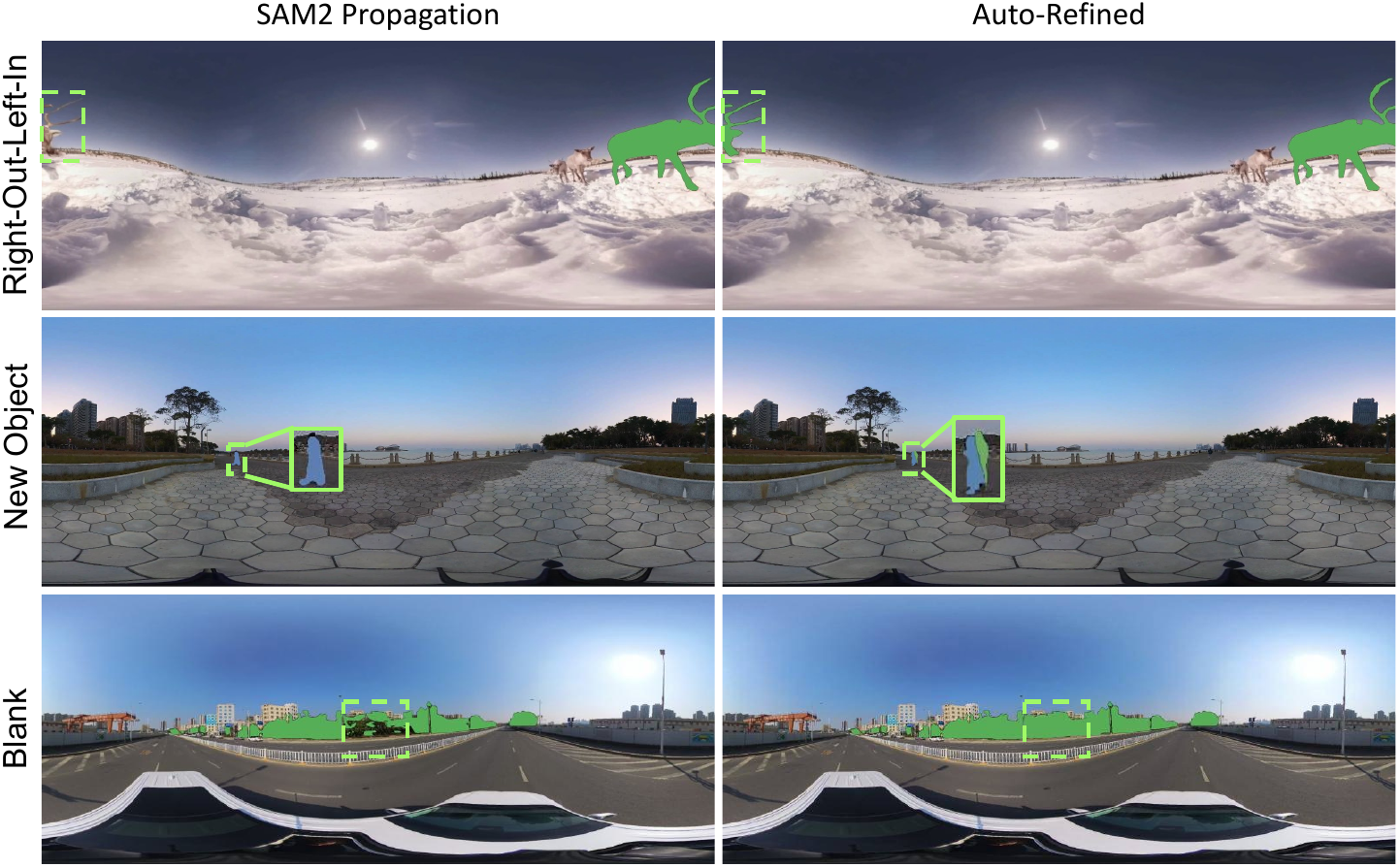}
  \end{center}
  \vspace{-10pt}
  \caption{\small{Example visualizations for the ablation of Auto-Refine Annotation Phase.}}
\vspace{-12pt}
\label{fig: ablation_auto_refine}
\end{minipage}
\begin{minipage}{0.45\linewidth}
\vspace{-5pt}
  \begin{center}
    \includegraphics[width=\linewidth]{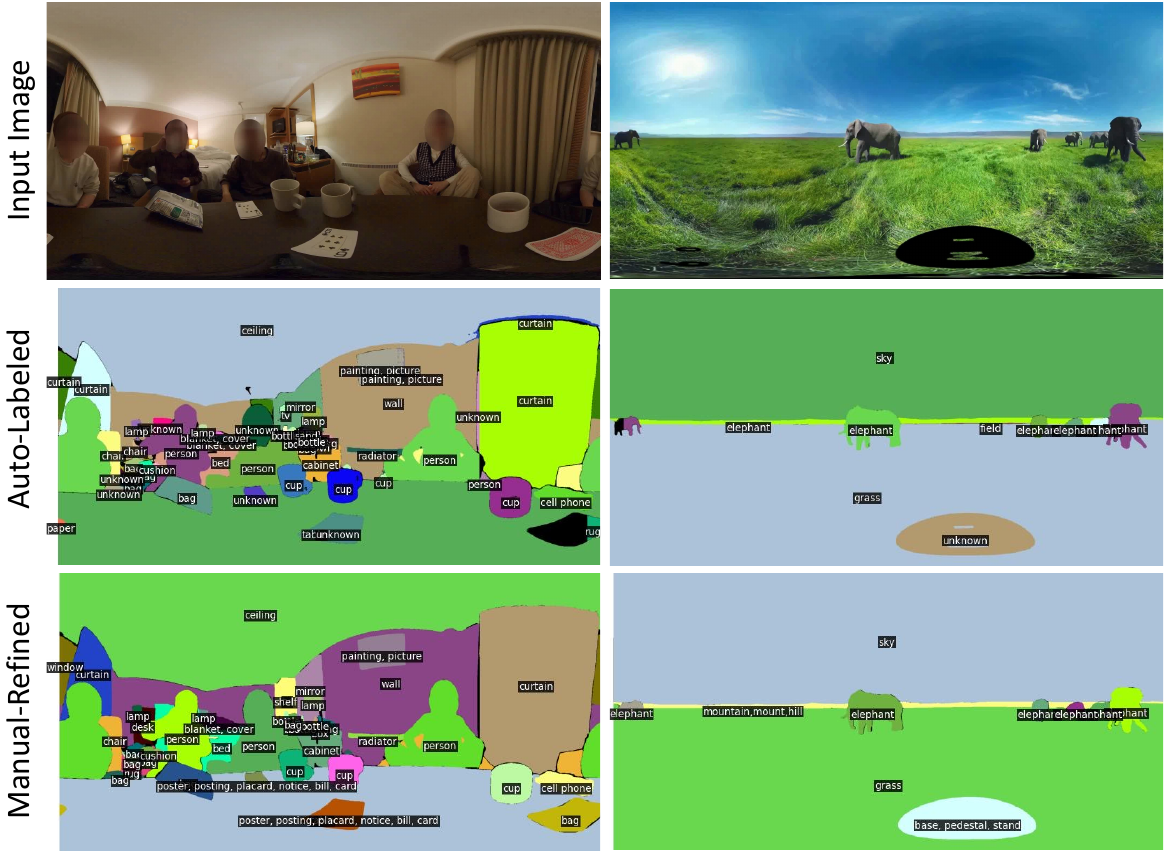}
  \end{center}
  \vspace{-10pt}
  \caption{\small{Example visualizations for the ablation of Manual Modification.}}
\label{fig: ablation_muanual_modification}
\end{minipage}
\vspace{-20pt}
\end{figure}

\noindent\textbf{Effectiveness of Phase II.}
% The results of the outputs from SAM2 and the auto-refinement process of Phase II, compared against the final annotations, are presented in Table \ref{table: component_analysis}. For updated frames, SAM2 results in large blank areas, which consequently leads to low $\mathcal{J} \& \mathcal{F}$ scores. However, following auto-refinement by Phase II, there is a notable increase of \textbf{+40.1} in terms of $\mathcal{J} \& \mathcal{F}$. This improvement is attributed to the refinement of existing masks and the addition of new masks facilitated by the MCR module. Examples of three cases involving blank areas are visualized in Fig.~\ref{fig: ablation_auto_refine}. The first case involves a deer that exits the right boundary of the frame and reenters from the left boundary, a scenario caused by the panorama effect. In Phase II's SDR Module, the left part of the deer is successfully segmented and assigned the same object ID as the right part, ensuring continuity. The second case illustrates a person who is obscured in the last frame but appears in the current frame. Here, the MCR Module segments the person and assigns a new label appropriately. The final case highlights a failure in SAM2's tracking, caused by panoramic distortion, which introduces a domain gap. The MCR Module corrects the segmentation error and restarts tracking at this frame, effectively restoring consistency.
Tab. \ref{table: component_analysis} presents a comparison of SAM2 outputs and Phase II’s auto-refinement process against final annotations. SAM2 results in large blank areas for updated frames, leading to low $\mathcal{J}\&\mathcal{F}$ scores. Phase II’s process increases these scores by \textbf{+40.1}, refining existing masks and adding new ones via the MCR Module. Fig.~\ref{fig: ablation_auto_refine} illustrates three cases. The first case involves a deer that exits the right boundary of the frame and reenters from the left boundary, a scenario caused by the panorama effect. In Phase II's SDR Module, the left part of the deer is successfully segmented and assigned the same object ID as the right part, ensuring continuity. The second case illustrates a person who is obscured in the last frame but appears in the current frame. Here, the MCR segments the person and assigns a new label appropriately. The final case highlights a failure in SAM2's tracking, caused by panoramic distortion, which introduces a domain gap. The MCR Module corrects the segmentation error and restarts tracking at this frame, effectively restoring consistency.

% a deer exits the right frame boundary and reenters from the left, successfully segmented by Phase II’s SDR module, maintaining the object ID; a person obscured in the last frame is labeled by the MCR Module upon appearance; and SAM2’s tracking failure due to panoramic distortion is corrected by the MCR Module, restarting tracking and restoring consistency.

\noindent\textbf{Effectiveness of Phase III.}
The Comparison between auto-refinement masks and manual-refined masks is shown in Fig.~\ref{fig: ablation_muanual_modification}. Although auto-refinement masks from our A$^3$360V pipeline demonstrate high quality, we still manually revise these masks to further enhance performance. During the revision, we specifically address issues related to object boundaries, label hallucinations, and incorrect masks.
% Specifically, we further adjust black edges between object boundaries generated by pre-trained 360-degree visual models and remove labels that are not included in our label vocabulary.

\subsection{Discussion}
\noindent\textbf{Flexibility of A$^3$360V.}
A$^3$360V's flexibility stems from its modular design, allowing users to select from various 2D segmentors for auto-annotation. For entity segmentation, options include SAM \cite{kirillov2023segment}, CropFormer \cite{qi2022high}, and E-SAM \cite{zhang2025sam}, providing robust object delineation. For panoptic segmentation, models like Mask2Former \cite{cheng2021mask2former, cheng2021maskformer}, OneFormer \cite{jain2023oneformer}, and OMG-Seg \cite{li2024omg} can be integrated for comprehensive scene understanding. A$^3$360V also supports the flexible selection of LLMs for label checking, ensuring compatibility with different user needs. This pipeline is versatile, applicable to both 360 and 2D videos, making it suitable for diverse video annotation tasks and adaptable to various datasets and applications. \textit{More discussions are in the Appendix.}

\section{Conclusion and Limitations}
\label{Conclusion}

\noindent \textbf{Conclusion.}
In this paper, we presented Leader360V, the first large-scale, labeled real-world 360 video dataset specifically designed for instance segmentation and tracking in diverse and dynamic environments. To reduce human labeling effort, we also proposed the A$^3$360V pipeline, a three-phase framework that integrated pre-trained 2D segmentors with large language models to automate the annotation process, with minimal human intervention limited to final refinement. Extensive user studies and experimental results demonstrated the effectiveness of each stage in the pipeline and highlight the potential of Leader360V to advance research in robust, scalable 360 video understanding.

\noindent \textbf{Broader Impacts.}
Leader360V has the potential to stimulate future research in 360VOT and 360VOS, and to support the development of foundation models tailored to 360 video understanding. Additionally, our A$^3$360V pipeline offers a practical paradigm for combining large language models with pre-trained vision models to reduce manual annotation costs, which may inspire more scalable and efficient dataset construction methods in future work.

\noindent \textbf{Limitations and Future Work.}
\label{sec:limitation}
Our current Leader360V dataset does not yet cover all object classes commonly encountered in daily life, nor does it include annotations for the motion states of moving objects. In future work, we plan to further enrich the dataset annotations to support a broader range of tasks. We also aim to leverage Leader360V to explore the development of foundation models for 360 visual understanding.

\clearpage

{
\small
\bibliographystyle{unsrt}
\bibliography{egbib}
}

% %%%%%%%%%%%%%%%%%%%%%%%%%%%%%%%%%%%%%%%%%%%%%%%%%%%%%%%%%%%%
\newpage
\appendix

\section{User Study}

\begin{wrapfigure}{r}{0.3\linewidth}
\vspace{-40pt}
% \centering
% \vspace{-10pt}
    \centering
    % \vspace{-25pt}
\includegraphics[width=0.9\linewidth]{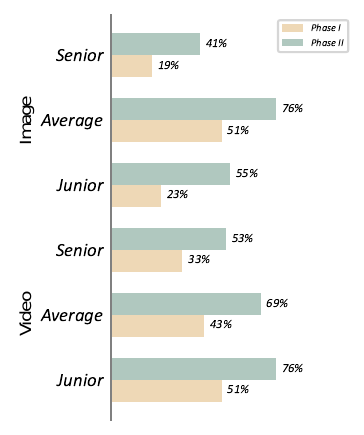}
% \caption{\small{The overall of our Leader360V dataset.}}
\vspace{-5pt}
\caption{{The user study of our Leader360V dataset.}}
\label{fig:distinguish}
\vspace{-15pt}
\end{wrapfigure}

To further investigate the effectiveness of our annotation pipeline, we randomly selected 100 videos (0.98\% of the entire dataset) from Leader360V, along with 500 images randomly selected from these videos. We invited three groups of human testers with junior, average, and senior skill levels in 360 video-related tasks, each consisting of 10 members.
In Fig.~\ref{fig:distinguish} illustrates testers' ability to distinguish between the manually revised image/video masks and the automatically annotated masks from Phase I (masks are produced by an open-source pre-trained model) or Phase II (masks are generated by our annotation pipeline, A$^3$360V). A higher score indicates that testers can more accurately identify differences.

We employed a 10-point scoring system for testers to evaluate these images and videos based on three metrics: the extent of mask missing (M), the extent of mask incorrectly annotated (W), and the annotation level on distorted objects (D). Detailed information is listed in the Table ~\ref{table:UserStudy}.
For \textbf{image evaluation}, a greater number of missing masks results in a lower M-score, and a higher number of incorrectly annotated masks leads to a lower W-score. Besides, the better the masks fit distorted objects, the higher the D-score.
For \textbf{video evaluation}, we focused on moving objects. A greater number of missing masks for moving objects results in a lower M-score, and a higher number of incorrectly annotated masks for newly appeared objects leads to a lower W-score. Besides, the better the masks fit distorted moving objects, the higher the D-score.

\begin{table*}[h!]
% \caption{\small{The user study of our Leader360V dataset.}}
    \renewcommand{\arraystretch}{1.2}
% \belowrulesep=0pt
% \aboverulesep=0pt
\centering
% {\caption
% {\textbf{User Study}}
% \label{table:UserStudyVideo}}
\caption{{The user study of our Leader360V dataset.}}
\label{table:UserStudy}
% \tiny
% \setlength{\tabcolsep}{0.7mm}
   \begin{tabular}{c|c|cccc|ccc|cccc}
        \toprule

        % \rowcolor{tablehighlightgray}
        \cellcolor{white}\multirow{3}{*}{\bf{Modality}} & \cellcolor{white}\multirow{3}{*}{\bf{Human}} & & \multicolumn{6}{c}{\textbf{Auto}} & & \multicolumn{3}{c}{\textbf{Revised}}   \\

        % \cline{2-10}
        \cdashline{4-9}
        \cdashline{11-13}
        % \cmidrule{10-12}
        % & \multicolumn{6}{c}{Auto} & \multicolumn{3}{c}{Human Revised}  \\

        & & & \multicolumn{3}{c|}{Phase I} & \multicolumn{3}{c|}{Phase II} & & \multicolumn{3}{c}{Phase III}  \\

        % \cline{2-4}
        % \cline{5-7}
        % \cline{8-10}

        % \rowcolor{tablehighlightgray}
        % \cellcolor{white} & \cellcolor{white} & \cellcolor{white} & Miss & Wrong & Distort &  Miss & Wrong & Distort & & Miss & Wrong & Distort  \\
        \cellcolor{white} & \cellcolor{white} & \cellcolor{white} & M & W & D &  M & W & D & & M & W & D  \\
        \midrule
        \midrule

        & Junior & & 7.9 & 7.2 & 8.4 &  8.4 & 7.9 & 9.0 & & 9.7 & 9.8 & 9.7  \\

        % \rowcolor{tablehighlightgray}
        \cellcolor{white}\textbf{Image} & Average & &  7.5 & 6.6 & 7.3  & 9.0 & 8.6 & 9.0 & & 9.3 & 9.9 & 9.7   \\

         & Senior &  & 8.0 & 5.7 & 7.3  & 9.1 & 7.7 & 8.9 & & 9.5 & 9.7 & 9.6 \\

        \midrule

        % \rowcolor{tablehighlightgray}
        \cellcolor{white}& Junior & & 5.3 & 5.1 & 6.4 &  7.4 & 7.0 & 7.9 & & 9.1 & 9.8 & 9.5    \\

        \textbf{Video} & Average & & 4.9 & 5.0 & 6.2 &  7.4 & 7.0 & 7.9 & & 9.1 & 9.8 & 9.5   \\

        % \rowcolor{tablehighlightgray}
        \cellcolor{white} & Senior & & 4.4 & 3.3 & 6.2 &  6.8 & 6.3 & 7.8 & & 8.9 & 9.6 & 9.2 \\

        \bottomrule
   \end{tabular}
% \end{minipage}
% \vspace{-20pt}
\end{table*}

\newpage
\section{Dataset}
\label{sec:dataset}
The videos in our Leader360V dataset are sourced from two main origins: existing 360 video datasets and newly collected videos from online resources or our own recordings.

\subsection{Videos From Existing Datasets}

As shown in Tab. \ref{table:datasource}, we manually filtered and selected videos from existing 360 video datasets originally collected for various 360-related tasks:
\begin{itemize}
    \item \textbf{360VOTS \cite{Xu2024360VOTSVO}} is a dedicated benchmark for 360 visual object tracking and segmentation, comprising 120 high-resolution video sequences totaling over 113K frames. It provides dense per-frame annotations and incorporates specialized representations—such as rotated bounding boxes and spherical region masks—to effectively handle challenges unique to omnidirectional content, including projection-induced distortion and left-right content discontinuities. The dataset is designed to support both tracking and segmentation tasks, with evaluation protocols and metrics tailored specifically for panoramic imagery.
    \item \textbf{PanoVOS \cite{Yan2023PanoVOSBN}} is a benchmark designed for video object segmentation in 360 videos. It consists of 150 panoramic video sequences annotated with over 19,000 instance masks, generated through a human-in-the-loop process that combines manual labeling with model-assisted propagation. The dataset features diverse real-world scenarios characterized by substantial camera motion and extended temporal duration, making it a valuable resource for evaluating long-term segmentation performance under the geometric challenges of panoramic video formats.

    \item \textbf{WEB360 \cite{wang2024360dvd}} is a dataset designed to support controllable 360 video generation, consisting of approximately 2,000 panoramic video-caption pairs collected from publicly available online sources. Each video is paired with a high-quality textual description, enabling text-conditioned 360° video synthesis. By facilitating generative modeling in the panoramic domain, WEB360 provides valuable training data for data-driven synthesis and serves as a foundation for expanding 360 video research beyond traditional perception tasks.

    \item \textbf{360+x \cite{chen2024360+}} is a large-scale multi-modal dataset curated for comprehensive panoramic scene understanding. It contains 2,152 videos recorded across 232 real-world environments, with each scene captured simultaneously using 360 cameras and egocentric (first-person) wearable devices. The dataset is further enriched with synchronized spatial audio, action annotations, GPS metadata, and scene-level context, supporting research on cross-view alignment, audio-visual reasoning, and multi-sensory perception in 360 video settings.

    \item \textbf{YouTube360 \cite{tan2024imagine360}} focuses on generating 360 videos from conventional narrow field-of-view (FOV) inputs. The dataset consists of over 10,000 equirectangular video clips, combining the WEB360 collection with more than 8,000 additional panoramas scraped from YouTube. These clips span a wide range of scenes, including urban environments, natural landscapes, and wildlife footage. YouTube360 enables learning-based view expansion and supports VR-oriented content creation, providing diverse training data for 360 video synthesis and immersive media generation.

\end{itemize}

\begin{table}[h]
    \centering
    % \vspace{-10pt}
    \caption{\small{\textbf{Our Data Source}.  \textbf{``Pct''}: percentage of selected data. \textbf{``Sel''}: specific number of selected data. VG: Video Generation. VC: Video Caption}}
   \label{table:datasource}
    % \vspace{-5pt}
    \small
    \setlength{\tabcolsep}{0.4mm}
   \begin{tabular}{rcccc}
        \toprule

        {\bf{Source*}} & {\bf{Task}} &  {\bf{Pct}} & {\bf{Sel}} & {\bf{Relabel}}\\
        % \cmidrule{2-4}
        \midrule

         360VOTS* \cite{Xu2024360VOTSVO} & 360VOT & 80\% & 232 & \textcolor{mygreen}{\ding{51}} \\
         PanoVOS* \cite{Yan2023PanoVOSBN} & 360VOS & 60\% & 90 & \textcolor{mygreen}{\ding{51}} \\
         WEB360* \cite{wang2024360dvd}  & 360 VG & 50\% & 1K+ & \textcolor{mygreen}{\ding{51}}\\
         360+x* \cite{chen2024360+} & 360 VC & 30\% & 1K+ & \textcolor{mygreen}{\ding{51}} \\
         YouTube360* \cite{tan2024imagine360} & 360 VC & 20\% & 3K+ & \textcolor{mygreen}{\ding{51}}\\
         Open Source & N/A & N/A & 1K+ & \textcolor{mygreen}{\ding{51}} \\
         Self-Collected & 360VOTS & N/A & 2K+ & \textcolor{mygreen}{\ding{51}}\\
        % 1057 646 1911
        \bottomrule
        % \vspace{-10pt}
   \end{tabular}
\end{table}

Since most of these datasets were not originally designed for 360 VOT or VOS tasks, we manually filtered motion-centric scenes to better align with our multi-task setting. During the selection process, we adhered to specific selection criteria: (1) Data involving scenes with dense objects, such as crowded people or stacked bicycles, were excluded, as such scenes are too complex even for SOTA methods. (2) Data that do not feature clearly moving objects, such as static footage captured in unchanging scenes, were not included. (3) Virtual videos constructed using virtual reality or simulators were also excluded. While these videos exist in existing datasets, they are challenging to classify as real-world scenarios.

In addition to those datasets not originally designed for 360 VOT or VOS tasks, as 360VOTS and PanoVOS only provide sparse annotations—typically focusing on a single target per video—we re-annotated all selected videos to include instance masks for all visible objects in each frame.

\begin{figure}[h!]
    \centering
    % \vspace{-25pt}
\includegraphics[width=1\linewidth]{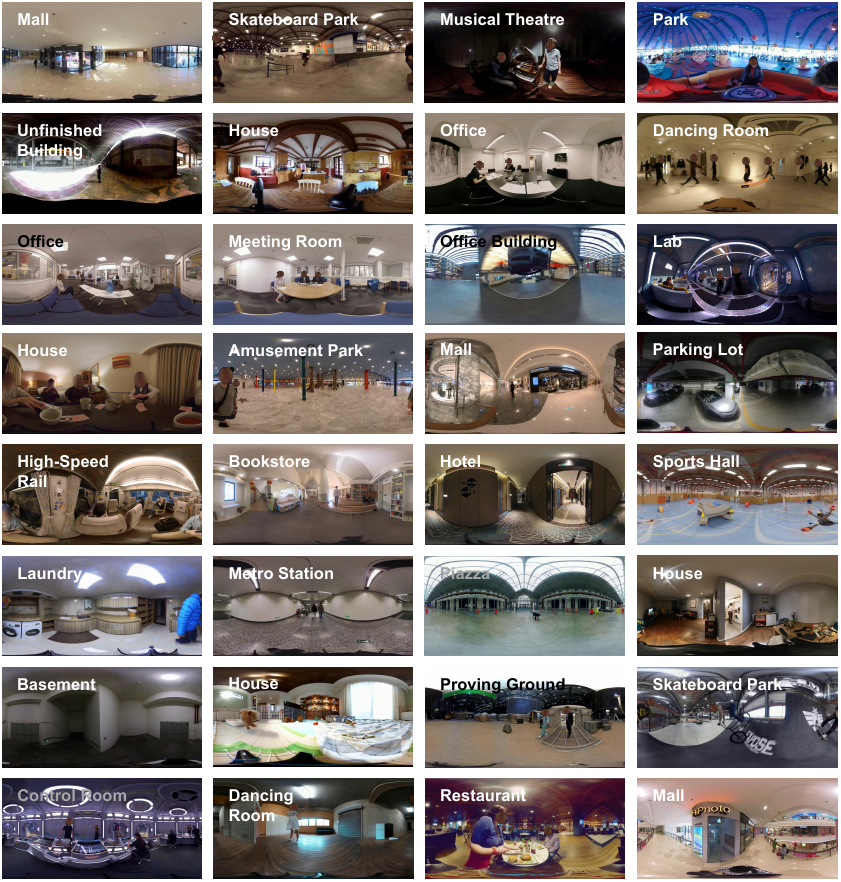}
\caption{\small{INDOOR Samples in Leader360V.}}
% \vspace{-10pt}
\label{fig: samples_indoor_supp}
\end{figure}

\begin{figure}[h]
    \centering
    % \vspace{-25pt}
\includegraphics[width=1\linewidth]{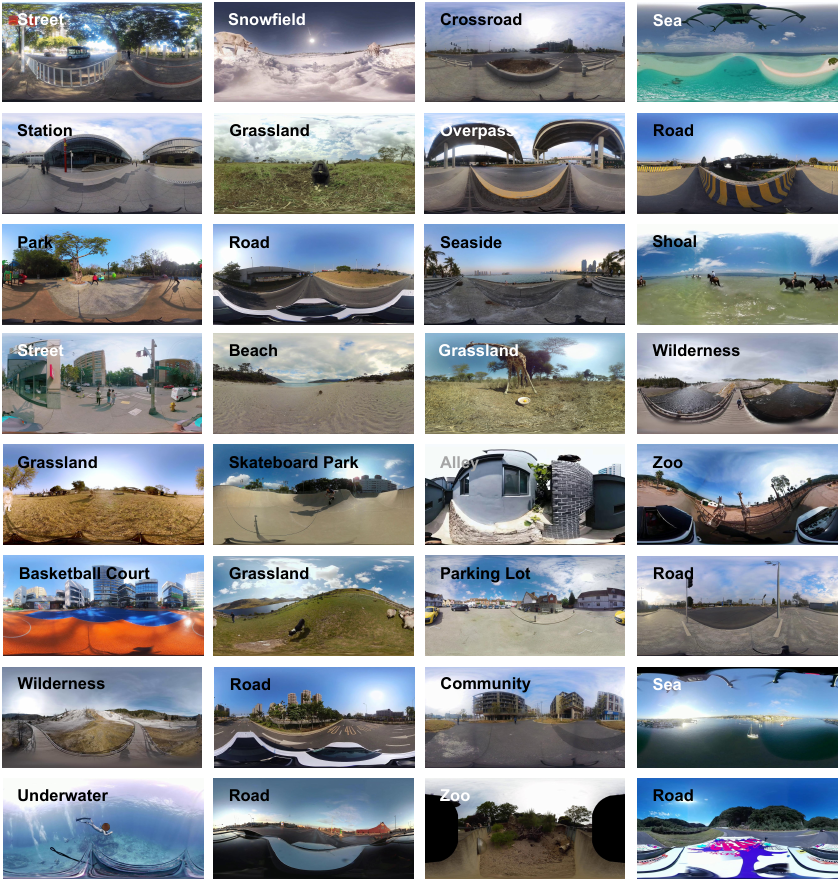}
\caption{\small{OUTDOOR Samples in Leader360V.}}
% \vspace{-10pt}
\label{fig: samples_outdoor_supp}
\end{figure}
\begin{figure}[h!]
    \centering
    % \vspace{-25pt}
\includegraphics[width=0.9\linewidth]{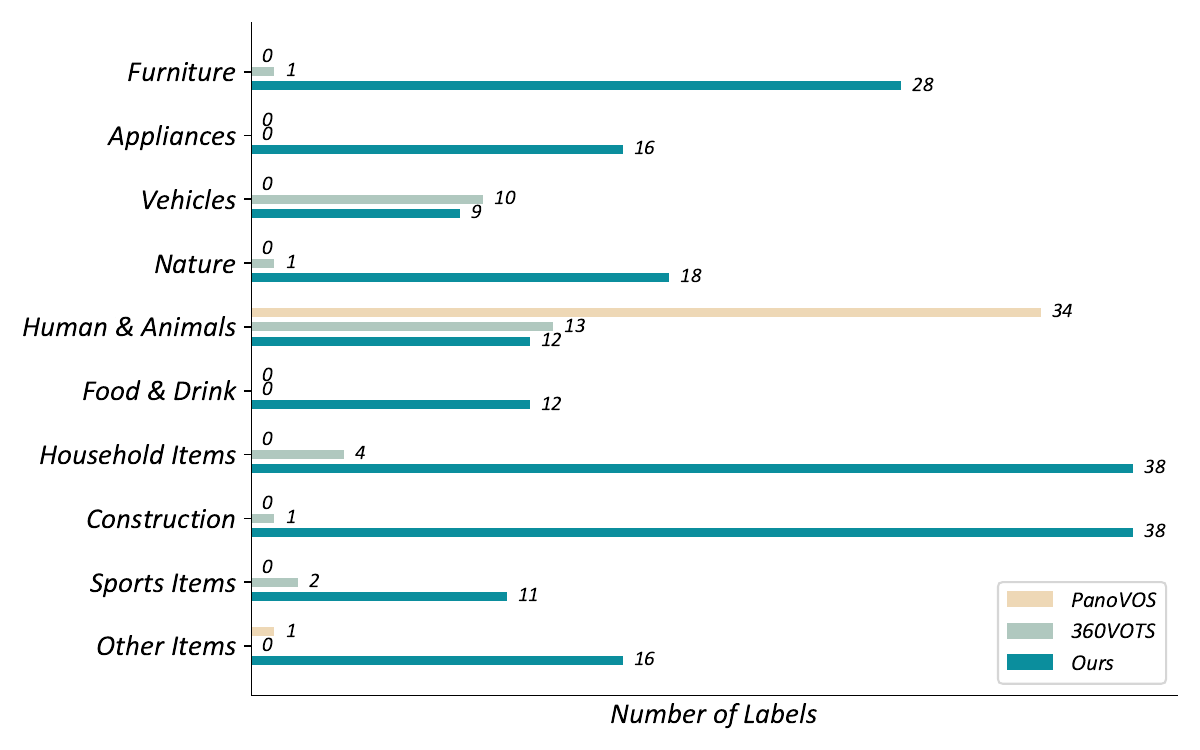}
\caption{\small{Comparison of category diversity among ours, PanoVOS, and 360VOTS}}
% \vspace{-10pt}
\label{fig: label_comparison_supp}
\end{figure}
\begin{figure}[h!]
    \centering
    % \vspace{-25pt}
\includegraphics[width=0.9\linewidth]{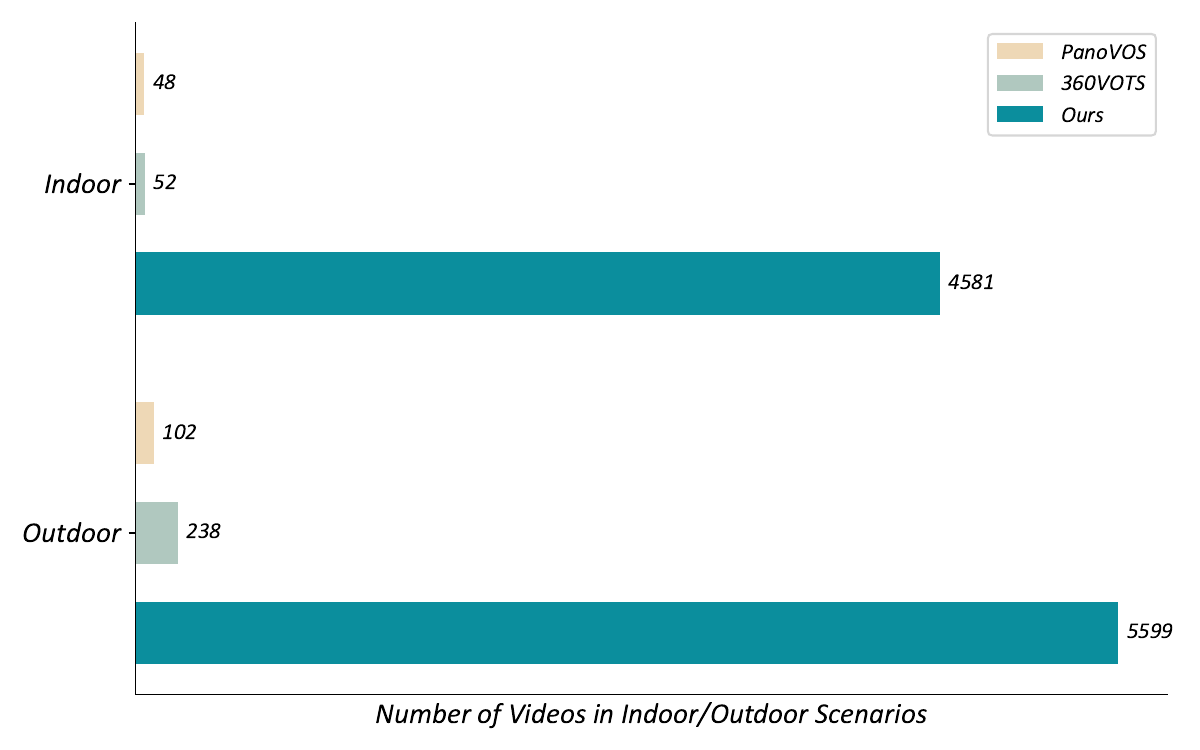}
\caption{\small{The overall of our Leader360V dataset.}}
% \vspace{-10pt}
\label{fig: scenario_comparison_supp}
\end{figure}

Moreover, the existing datasets exhibit imbalances in scene diversity, as shown in Fig. \ref{fig: label_comparison_supp} and Fig. \ref{fig: scenario_comparison_supp}. This is evident in three key aspects: \textbf{first}, there is a significant imbalance in the ratio of indoor to outdoor videos within the dataset, as well as in the distribution of various types of labels; \textbf{second}, the ratio of dynamically shot videos to stationary ones is unbalanced; \textbf{third}, among the dynamically shot videos, the ratio of videos featuring human subjects versus those featuring vehicles (including cars, drones, etc.) is also significantly skewed. By considering the varying speeds of each vehicle, we aim to enrich the dataset's diversity, thereby mitigating the risk of overfitting. Specifically, PanoVOS \cite{Yan2023PanoVOSBN} and 360VOTS \cite{Xu2024360VOTSVO} exclude labels for categories other than animals and vehicles, resulting in an uneven and unreasonable distribution of labels. Additionally, the number of outdoor videos is several times greater than that of indoor videos. However, our dataset addresses these imbalances by deliberately controlling the number of videos for each scenario.

% \textcolor{red}{More details about filter guideline, more description about scene limitation/ capture limitation(static dynamic)}

% \begin{table}[h]
% \centering
% \caption{Diverse360: Indoor and Outdoor scenes and Image Counts}
% \begin{tabular}{@{}clp{10cm}@{}}
% \toprule
% \multirow{2}{*}{\textbf{Scenes}} & \textbf{Indoor} & Classrooms, libraries, corridors, meeting rooms, gymnasiums \\ \cmidrule{2-3}
%                                     & \textbf{Outdoor} & Outside dormitories, outside office buildings, outside gymnasiums, outside activity centers \\ \midrule
% \multirow{2}{*}{\textbf{\# Images}} & \textbf{Indoor} & 7,887 \\ \cmidrule{2-3}
%                                     & \textbf{Outdoor} & 4,176 \\ \bottomrule
% \end{tabular}
% \label{tab:datasets}
% \end{table}

\subsection{Videos From Self-collected}
As the videos selected from existing datasets lacked sufficient coverage of complex urban and indoor environments, we additionally collected new videos through our own recordings and publicly available online sources.

% \textcolor{red}{More details about 360 camera, FPS, Resolution}
\noindent\textbf{Our new recordings.} During the data collection, we use the \textit{Insta360 X3} camera to capture videos with at a resolution of 2048 $\times$ 1024, with a frame rate of 30 frames per second. To ensure rich scene diversity, we recorded new videos across more than ten cities, covering a wide range of both indoor and outdoor environments. Some samples from our dataset are listed in Fig. \ref{fig: samples_indoor_supp} and Fig. \ref{fig: samples_outdoor_supp}. Various capturing strategies were employed, including tripod-mounted stationary recording, handheld shooting with a selfie stick for walk-through scenes, and vehicle-mounted setups to simulate autonomous driving scenarios. In addition, we captured multiple views of the same scene from different angles to enhance spatial coverage.

\noindent\textbf{Online Sources.} To construct the first large-scale 360 video dataset, we additionally curated a set of newly uploaded high-resolution 360 videos from online platforms. The collection process followed four key criteria: (1) no violation of creator privacy or regional bias; (2) presence of clear object motion; (3) resolution no lower than 480p; and (4) absence of severe illumination issues, occlusions, or compression artifacts. After applying face anonymization to protect privacy, the collected videos were temporally segmented into clips with an average duration of approximately 15 seconds. In total, we successfully gathered over 1K video sequences from multiple online sources.

\newpage
\section{More Discussions}
\subsection{Bounding Box Annotation.}
\begin{wrapfigure}{r}{0.45\linewidth}
\vspace{-20pt}
  \begin{center}
    \includegraphics[width=0.45\textwidth]{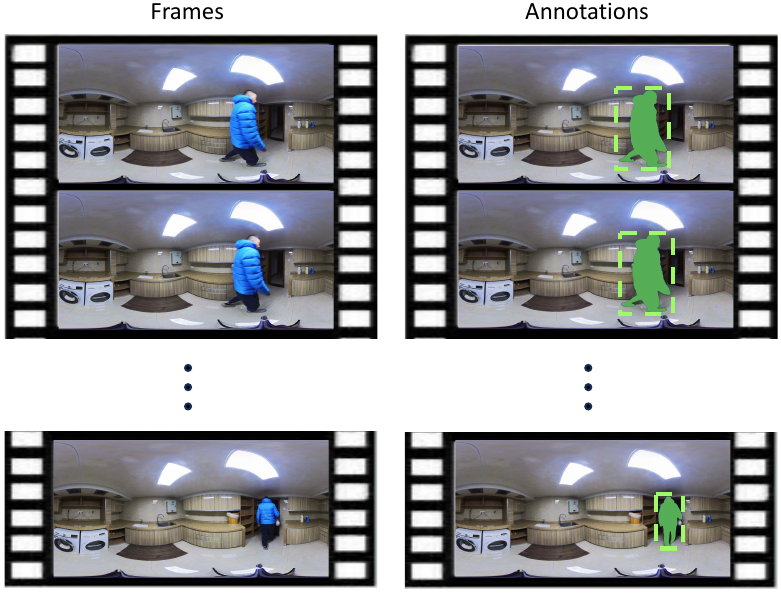}
  \end{center}
  \vspace{-5pt}
  \caption{\small{Samples of the bounding box annotation for VOT in the Leader360V dataset.}}
\vspace{-20pt}
\label{fig: tracking_sample}
\end{wrapfigure}
In addition to mask annotations, the Leader360V dataset includes bounding box annotations, which play a critical role in supporting various computer vision tasks such as object detection and tracking.
Fig.~\ref{fig: tracking_sample} demonstrates a sample image from the dataset, highlighting how bounding box annotations accurately localize objects. The inclusion of these detailed annotations not only complements the mask annotations but also significantly broadens the dataset's applicability. This makes Leader360V a comprehensive and versatile resource, enabling researchers and developers to tackle a wide range of challenges in computer vision, particularly for 360 image and video analysis.

\subsection{Origin of Semantic Labels}
Our semantic labels are derived from COCO\cite{lin2014microsoft}, ADE20K\cite{zhou2019semantic}, and Cityscapes\cite{Cordts2016Cityscapes}. During the merging process, we encountered conflicts and duplications among labels, which required careful resolution. To address this, we adhered to three key rules:
\begin{itemize}
    \item Labels with similar or identical semantics across datasets were merged into a single unified category to ensure consistency (e.g., \textit{building-other-merged} from COCO\cite{lin2014microsoft} was unified into the broader \textit{building} category).
    \item Unique labels appearing in only one dataset but distinctly different from others were retained (e.g., \textit{fountain} from ADE20K\cite{zhou2019semantic}).
    \item Rare or ambiguous subclasses (e.g., \textit{rider}) were either removed or merged into broader categories like \textit{person}.
\end{itemize}
Throughout this process, we prioritized maintaining the diversity of labels to ensure that the dataset remains comprehensive and effective for a wide range of segmentation tasks.

\subsection{Analysis of Annotation Cost}
\label{resource_cost}
Annotation is an essential yet highly time-consuming process for creating pixel-level segmentation datasets. For full manual annotation, processing a single video at 1 fps can take over 20 hours. However, with our innovative A$^3$ 360V pipeline, this time is significantly reduced to just 2 hours for the same frequency, greatly improving efficiency. For automatic annotation, we utilize eight powerful 80GB H100 GPUs, which enable us to annotate 250 videos in just 2 days. The pipeline is capable of simultaneously tracking up to 63 objects, ensuring robust multi-object segmentation. This streamlined approach saves considerable time while maintaining high-quality results.

\subsection{Flexibility of A$^3$360V.}
A key strength of A$^3$360V lies in its high flexibility and modular design, which allows users to freely customize components based on specific annotation needs and target tasks. The framework supports interchangeable choices of 2D segmentors, large language models (LLMs), and downstream prompt strategies, enabling seamless adaptation across diverse video annotation scenarios.
For 2D segmentation backbones, A$^3$360V offers configurable support for both entity-level and panoptic-level models. For entity segmentation, users may select from models such as SAM~\cite{kirillov2023segment}, CropFormer~\cite{qi2022high}, and E-SAM~\cite{zhang2025sam}, each offering strong object delineation capabilities. For panoptic segmentation, our framework integrates models like Mask2Former~\cite{cheng2021mask2former, cheng2021maskformer}, OneFormer~\cite{jain2023oneformer}, and OMG-Seg~\cite{li2024omg}, each of which supports multiple semantic class taxonomies (e.g., COCO, ADE20K, Cityscapes), thereby enriching the semantic diversity and label granularity available to users.
In addition, A$^3$360V supports flexible integration of LLMs for semantic label verification. The LLM-based agent can be customized via task-specific text prompts, allowing users to repurpose the same module for different roles such as label disambiguation, object categorization, or hierarchical grouping. This prompt-driven flexibility also opens opportunities for expanding A$^3$360V to other 360 video understanding tasks, such as video captioning, by simply replacing the 2D segmentors and reconfiguring the LLM agent's prompt structure.
Notably, the proposed framework is not limited to 360 content. With minimal adjustments—e.g., removing ERP-specific pre-processing steps—A$^3$360V can be directly applied to large-scale 2D video datasets, enabling multi-task annotation with minimal human intervention. This generalizability makes A$^3$360V a practical and extensible tool for a wide range of video annotation applications.

\subsection{Category Superiority and Constraints.}
As illustrated in Fig.~\ref{fig: label_comparison_supp} and Fig.~\ref{fig: scenario_comparison_supp}, our Leader360V dataset encompasses a significantly larger number of semantic categories compared to existing 360VOT and 360VOS datasets. This highlights the rich semantic diversity of our annotations and the potential value of Leader360V for training more practical and generalizable 360 foundation models. While the high category count benefits from the prior knowledge embedded in pre-trained 2D segmentors and the recognition capabilities of large language models (LLMs), we do not pursue label diversity for its own sake. Excessive granularity in class definitions can lead to a highly imbalanced category distribution, where certain niche classes are represented in only a few isolated videos.
To address this issue, we adopt a class merging strategy to ensure semantic coherence while preserving practical utility. For instance, in the 360+x dataset, a variety of animal-related categories are present—such as "penguin" or "gorilla"—which are rarely seen in either our newly collected videos or other public 360 datasets. As a result, we group such infrequent categories under a generalized label “other animals,” while retaining commonly observed classes like “cat” and “dog” as distinct entries in our final taxonomy.
Additionally, we unify synonymous or hierarchically similar categories across different segmentation taxonomies. For example, “airplane” and “plane” are merged under the label “plane”; fine-grained wall types from ADE20K such as “wall-brick,” “wall-stone,” “wall-tile,” “wall-wood,” and “wall-other-merged” are all consolidated under the label “wall.” Throughout the LLM-guided annotation and final review process, we also maintain a record of novel labels proposed by the LLM agent. These candidate labels are manually reviewed and filtered before being included in the final label set.
Through this class consolidation process, we arrive at a carefully curated taxonomy consisting of 198 semantic categories in Leader360V. This design balances label richness with distributional consistency, supporting both detailed scene understanding and stable model training across diverse 360 video scenarios.

\subsection{The Value of Leader360V.}
As the first large-scale, real-world 360 video dataset with rich scene diversity and semantic coverage, Leader360V provides significant value for both core tasks in the 360VOTS benchmark—visual object segmentation and tracking. It addresses the current lack of well-annotated large-scale datasets in the 360 domain, which has been a major bottleneck for developing and evaluating robust models. By offering dense, high-quality annotations across a wide range of environments and object categories, Leader360V helps overcome the limitations of existing methods in semantic recognition and scene generalization and is expected to catalyze future research progress in both 360 segmentation and tracking.
Beyond task-specific benchmarks, Leader360V also fills a critical gap in the 360 community: the absence of a comprehensive dataset to support the training of large-scale segmentation and tracking foundation models. Existing approaches often rely on stitching together outputs from multiple 2D models deployed across different camera views, which incurs high computational costs and fails to capture the holistic structure of spherical scenes. In contrast, Leader360V provides a unified, native 360 representation that can support end-to-end training of 360 vision foundation models, thereby offering a scalable and efficient alternative. We believe this dataset will contribute meaningfully to the broader development of 360 video understanding in both academic and real-world applications.
\begin{figure}[t!]
    \centering
\includegraphics[width=\linewidth]{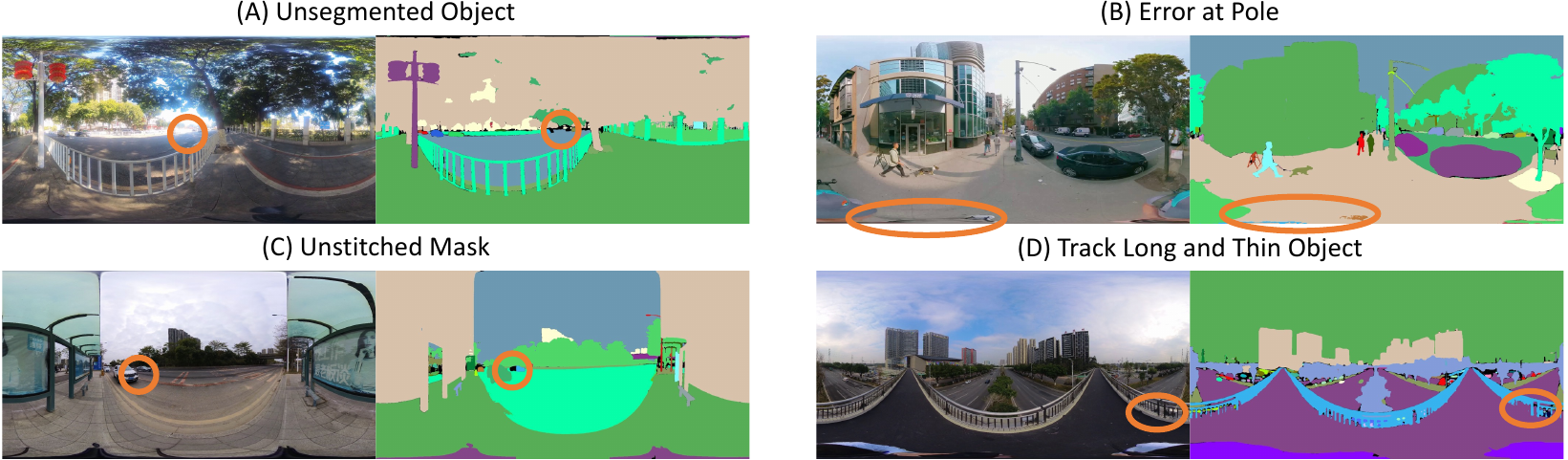}
\caption{\small{Visualization of A$^3$360V failure cases. Case \textit{A} shows the unsegmented part. Case \textit{B} is the wrong segmentation near the pole. Case \textit{C} visualizes the failure case of window stitching. Case \textit{D} shows disability of the employed MOT model to track long and thin objects (like rail).}}
% \vspace{-10pt}
\label{fig: failure_cases}
\end{figure}

\subsection{Failture Cases of A$^3$360V.}
In Fig.~\ref{fig: failure_cases}, notable failure cases of A$^3$360V auto annotation pipeline are presented, underscoring challenges specific to 360 video. In the first case, the system fails to segment objects near the equator, as they are not obvious to the entity segmentation model. The second case illustrates issues occurring near the zenith or nadir of 360 images, where distortions can lead to significant segmentation errors, revealing the pipeline's limitations in handling objects around the pole. The third case is inaccurately stitching together segmented regions across the full panoramic view, causing misalignments that disrupt the coherence of the scene. The last case demonstrates the system's struggle with maintaining consistent tracking of elongated, thin objects like railings, often resulting in fragmented or missing annotations, due to the limitation of the employed MOT model (SAM2\cite{Ravi2024SAM2S}). These challenges highlight the necessity for more advanced algorithms that can effectively handle the unique spatial and geometric complexities of 360 video environments, enabling more robust and accurate annotations. In future work, we plan to explore additional strategies to further enhance the performance and adaptability of our A$^3$360V framework.

\newpage

\end{document}